\documentclass[11pt]{article}
\usepackage[preprint]{acl}
\PassOptionsToPackage{hyperfootnotes=false}{hyperref}

\usepackage{times}
\usepackage{latexsym}
\usepackage{amsmath}
\usepackage{amssymb}
\usepackage{booktabs} 
\usepackage{multirow} 
\usepackage[T1]{fontenc}
\usepackage[utf8]{inputenc}

\usepackage{array}
\usepackage{microtype}

\IfFileExists{inconsolata.sty}{\usepackage{inconsolata}}{}

\usepackage{graphicx}
\usepackage{enumitem}
\usepackage{pgfplots}
\usepackage[caption=false,font=small]{subfig}
\pgfplotsset{compat=1.16}
\definecolor{taldblue}{HTML}{245F9E}
\definecolor{taldmidblue}{HTML}{5096DE}
\definecolor{taldlightblue}{HTML}{A6C8EA}
\definecolor{taldgray}{HTML}{897675}
\definecolor{taldred}{HTML}{B92E26}
\definecolor{taldaxis}{HTML}{707780}
\definecolor{taldgrid}{HTML}{DADDE1}
\pgfplotsset{
    tald plot/.style={
        axis line style={taldaxis, line width=0.6pt},
        tick style={taldaxis, line width=0.6pt},
        tick align=outside,
        tick label style={font=\small\sffamily, text=black},
        label style={font=\small\sffamily, text=black},
        title style={font=\small\sffamily\bfseries, text=black},
        ymajorgrids=true,
        grid style={taldgrid, line width=0.5pt},
        legend style={draw=none, fill=none, font=\scriptsize\sffamily},
    },
}

\title{When Do LLMs Apply the Wrong Law?\\
Diagnosing LLM Failures in Temporal Legal Reasoning}

\author{
\textbf{Yiqian Huang\textsuperscript{1}%
  },
  \textbf{Yang Zhou\textsuperscript{4}},
  \textbf{Qianying Liu\textsuperscript{3}},
    \textbf{Shaowen Peng\textsuperscript{5},}
    \textbf{Yuntao Kong\textsuperscript{6}},
\\
  \textbf{Kotaro Funakoshi\textsuperscript{1}},
  \textbf{Chuan Xiao\textsuperscript{2},}
  \textbf{Manabu Okumura\textsuperscript{1},}
  \textbf{Yang Cao\textsuperscript{1}}
  \textbf{Shuyuan Zheng\textsuperscript{2}%
\thanks{Corresponding author.}},
\\
  \textsuperscript{1}Institute of Science Tokyo
  \textsuperscript{2}Osaka University
  \textsuperscript{3}NII LLMC
  \textsuperscript{4}University of Tokyo
\\
  \textsuperscript{5}Nara Institute of Science and Technology
  \textsuperscript{6}Center of Juris-Informatics, ROIS-DS
\\  \texttt{h1k@lr.first.iir.isct.ac.jp, zheng@ist.osaka-u.ac.jp} 
}

\usepackage{placeins}

\input{savespace.tex}
\makeatother
\usepackage{enumitem}
\setlist[itemize]{leftmargin=0pt, itemindent=0pt, labelsep=0.5em}
\setlist[enumerate]{leftmargin=0pt, itemindent=0pt, labelsep=0.5em}

\SetEnumitemKey{questionnaireoptions}{labelindent=0.5em, labelwidth=0.5em,
  leftmargin=!, itemindent=0pt, labelsep=0.5em, align=left}


\usepackage{soul}
\usepackage{tcolorbox}
\tcbuselibrary{skins,breakable}

\newcommand{\myparagraph}[1]{\par\noindent\textbf{#1}}

\definecolor{lightred}{RGB}{255, 200, 200}
\definecolor{lightgreen}{RGB}{200, 255, 200}
\definecolor{deepviolet}{RGB}{96, 48, 96}

\tcbset{
  promptbox/.style={
    breakable,
    top=10pt,
    colback=white,
    colframe=violet,
    colbacktitle=deepviolet,
    fontupper=\itshape,
    enhanced,
    center,
    attach boxed title to top left={yshift=-0.1in,xshift=0.15in},
    boxed title style={boxrule=0pt,colframe=white},
  }
}
\newtcolorbox{promptframed}[2][]{promptbox,title=#2,#1}

\tcbset{
  promptbox1/.style={
    colback=white,
    colframe=violet,
    fontupper=\itshape,
    enhanced,
    center,
  }
}
\newtcolorbox{promptframed1}[2][]{promptbox1,title=#2,#1}

\newcounter{promptcounter}
\newenvironment{prompt}[1]
{
  \refstepcounter{promptcounter}\par\medskip
  \begin{promptframed}{Prompt~\thepromptcounter: #1}
}
{
  \end{promptframed}
}

\begin{document}
\maketitle

\begin{abstract}
Legal reasoning requires identifying which version of a statute applies to a case, typically based on when the legally relevant events occurred rather than when the case is adjudicated. Yet whether large language models (LLMs) can reliably perform such temporal legal reasoning remains largely unexplored. We formalize this task as Temporal Applicable-Law Determination (TALD), and construct a TALD-Bench, a benchmark of 51,956 Chinese civil judgments spanning the enactment of the Civil Code. Evaluating nine advanced LLMs, we identify three key findings. First, models exhibit a strong bias toward newer law, achieving over 70\% accuracy on new-law cases but below 25\% on old-law cases. Second, models demonstrate substantial knowledge of statutory versions and temporal-effect rules, yet struggle to apply this knowledge to cases. Stronger general reasoning performance does not consistently translate into better TALD accuracy, whereas temporal-law hints yield substantial improvements, suggesting a bottleneck in how models deploy their existing capabilities. Third, temporal-law hints also increase uncertainty during statutory-version reasoning, suggesting that they may help models reconsider their default preference for newer law. Our findings highlight the importance of adapting how LLMs apply existing capabilities, rather than relying solely on improvements in legal knowledge or general reasoning.\footnote{Our code and data will be released upon acceptance.}
\end{abstract}

% \section{Introduction}
\section{Introduction}

\begin{figure*}[t]
\centering
\includegraphics[width=0.9\textwidth]{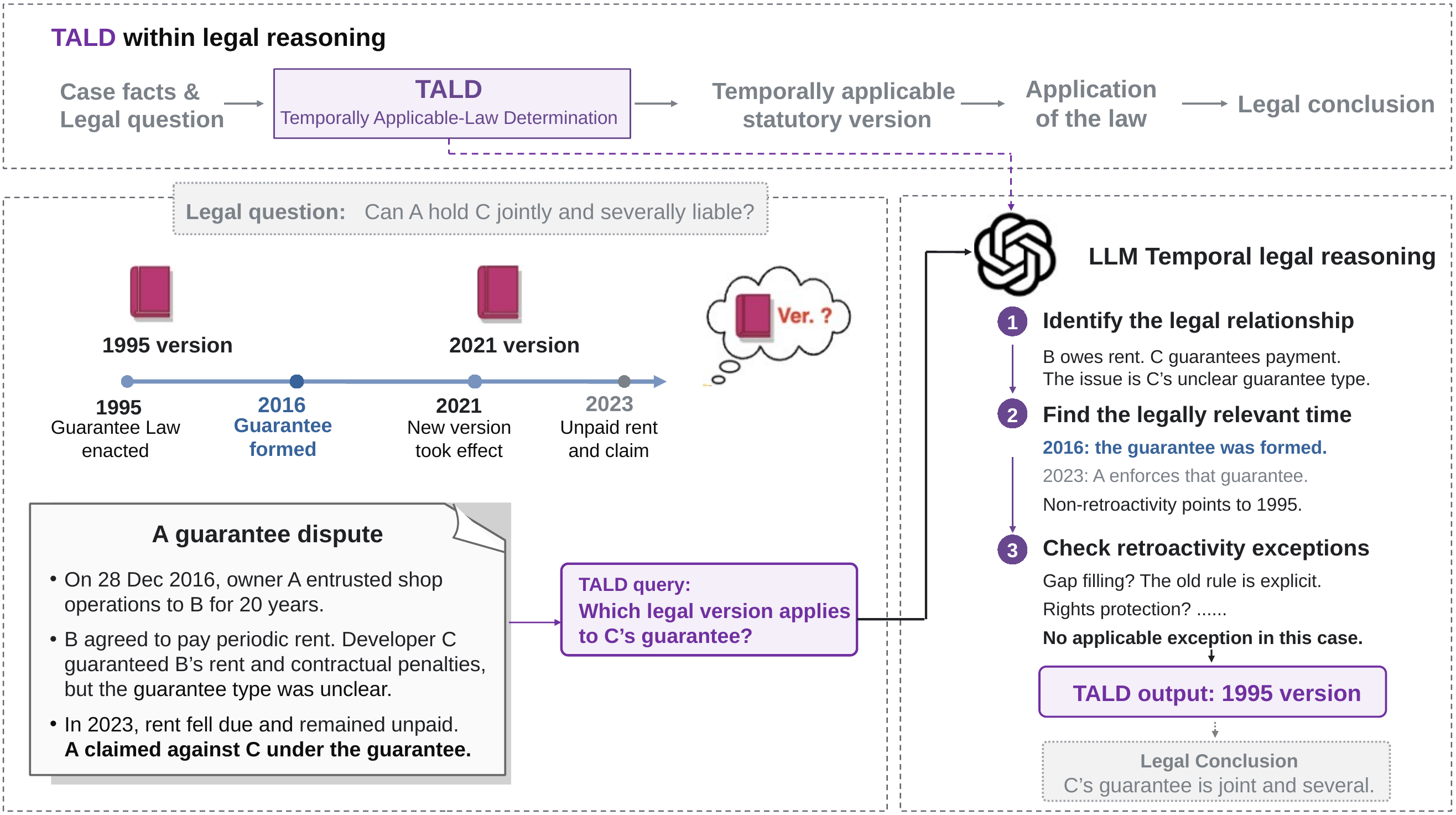}
\caption{Temporal Applicable-law Determination (TALD) as a vital step within legal reasoning, and an illustrative instance of how temporal legal reasoning is implemented by LLMs when faced with a legal issue.}
\label{fig:intro_illustration}
\end{figure*}

Large language models are increasingly applied to legal tasks~\citep{cui2023chatlaw,huang2023lawyer} such as legal judgment prediction (LJP)~\citep{luo2017learning}, legal fact prediction~\citep{liu2025legal}, court opinion generation~\citep{li2021court}, and contract analysis~\citep{hendrycks2021cuadexpertannotatednlpdataset}. However, reliable legal reasoning requires more than identifying relevant legal provisions: \textit{models must also determine which version of a law applies to the case at hand.}  A critical yet underexplored dimension of legal reasoning is its inherently \emph{temporal} nature. As laws are enacted and amended over time, different versions of the same statute may govern otherwise similar disputes. Under the principle of non-retroactivity, the applicable law is generally determined by when the legally relevant events occurred, subject to legally defined exceptions \cite{bowen2005retroactivity}. 
This becomes particularly challenging when a case involves events spanning multiple statutory versions, as illustrated in Figure~\ref{fig:intro_illustration}. For example, a dispute adjudicated in 2023 may concern a guarantee established in 2016, before the enactment of a new statutory framework in 2021. Correctly identifying the governing law therefore requires determining which events are legally decisive and how temporal-effect rules apply to them. We refer to this capability as \textbf{temporal legal reasoning}.

Despite its importance, temporal legal reasoning remains underexplored in existing LLM evaluations. Previous studies have examined temporal aspects of legal analysis, including event ordering, temporal distribution shifts, and adaptation to statutory revisions~\citep{barale2025lextime, santosh2025lextempus, han2026lawshift}.
However, these studies do not systematically evaluate whether LLMs can identify the legally applicable statutory version from the facts of a case. This distinction is important: knowing which laws exist at different points in time does not necessarily imply an ability to determine which version governs a particular dispute.
To investigate this capability, we introduce \textbf{T}emporal \textbf{A}pplicable-\textbf{L}aw \textbf{D}etermination (\textbf{TALD}), a task that requires models to identify the applicable statutory version based on case facts and legal questions. We construct a benchmark comprising 51,956 real Chinese civil judgments spanning the enactment of the Chinese Civil Code, with reference labels derived from the statutory provisions cited in judicial decisions. This benchmark enables us to systematically evaluate temporal version selection and investigate the sources of LLM failures.

Our evaluation of nine advanced LLMs investigates why they struggle and how targeted guidance affects their reasoning, resulting in three key findings. 
First, models exhibit a strong preference for newer law, achieving over 70\% TALD scores on new-law cases but below 25\% on old-law cases. Error analysis confirms that models frequently apply newer provisions when earlier laws govern. Second, models perform well on legal knowledge probes, yet struggle to apply this knowledge to individual cases. General reasoning performance does not consistently predict TALD performance, whereas simple temporal-law hints substantially improve it. This contrast suggests a gap between models' demonstrated capabilities and their application to temporal legal reasoning. Third, temporal-law hints improve accuracy while increasing uncertainty during statutory-version reasoning, suggesting that they may encourage models to consider alternatives rather than immediately favoring newer law.

Our contributions are threefold:
\begin{itemize}[leftmargin=*, nosep]
    \item \textbf{Task and benchmark.} We introduce TALD and construct TALD-Bench, a large-scale, judgment-derived benchmark for evaluating LLMs' ability to select the temporally applicable statutory version.
    \item \textbf{Systematic diagnosis.} We identify a strong newer-law bias and reveal a substantial gap between models' demonstrated legal knowledge and general reasoning performance and their ability to apply these capabilities to temporal legal reasoning. 
    \item \textbf{Behavioral analysis.} We demonstrate that temporal-law guidance substantially improves TALD performance and provide token-level evidence consistent with models reconsidering statutory alternatives, motivating further investigation into how domain-specific reasoning behavior can be improved.
\end{itemize}
% \textbf{(1) Task and benchmark.} We introduce TALD and construct TALD-Bench, a large-scale, judgment-derived benchmark for evaluating LLMs' ability to select the temporally applicable statutory version. \textbf{(2) Systematic diagnosis.} We identify a strong newer-law bias and reveal a substantial gap between models' demonstrated legal knowledge and general reasoning performance and their ability to apply these capabilities to temporal legal reasoning. \textbf{(3) Behavioral analysis.} We demonstrate that temporal-law guidance substantially improves TALD performance and provide token-level evidence consistent with models reconsidering statutory alternatives, motivating further investigation into how domain-specific reasoning behavior can be improved.

\section{The Temporal Applicable-Law Determination Task}
\label{sec:task_dataset}

\subsection{Preliminaries}

Statutes evolve through enactment, amendment, and repeal, creating multiple versions of provisions governing the same legal issue. Under the principle of \textit{non-retroactivity}, the applicable version is generally determined by when the legally relevant events occurred. Exceptions may nevertheless permit newer provisions to apply to earlier events, for example, to fill gaps in earlier law or better protect parties' rights~\citep{kryvoi2021non,spc2020temporal}.

Applying these principles requires identifying the legally decisive events and assessing their relationship to statutory changes. As illustrated in Figure~\ref{fig:intro_illustration}, a dispute arising after a statutory revision may still be governed by earlier law. Temporal legal reasoning therefore involves more than recalling statutory provisions or their effective dates: models must determine which version governs the specific legal issue. This determination provides the legal basis for subsequent reasoning, as different versions may assign different consequences to the same facts.

% Statutes are enacted, amended, and repealed over time, creating different versions of the rules governing the same legal issue.
% Under the principle of \textit{non-retroactivity}, the applicable version is generally the one in force when the legally relevant facts occurred.
% Exceptions may nevertheless allow newer provisions to apply to earlier legal facts, for example, to better protect parties' rights or fill gaps in earlier law~\citep{kryvoi2021non,spc2020temporal}.

% Determining the applicable version therefore requires \textit{temporal legal reasoning}.
% A case may involve several events spanning a statutory change, and the legal issue determines which events and dates matter.
% Models must reconstruct the relevant timeline, interpret the relationships among the facts under the relevant statutory provisions, and assess whether an exception changes which version applies.
% This requires applying knowledge of both statutory content and temporal-effect rules to the specific facts of a case.
% The selected version establishes the legal basis for subsequent reasoning, since different versions may assign different legal consequences to the same facts.
 
\subsection{Task Definition}
\label{ssec:task_definition}
% Consider that we have a collection of statutes $N = \{1, \dots, n\}$, where each statute $i \in N$ may have multiple versions due to temporal legal evolution.
% % Let a random variable $Y_i$ denote the specific version of statute $i$ to be cited in a legal reasoning process, where $Y_i = -1$ means no versions of statute $i$ are cited. 
% Given a query $(F, Q)$, where $F$ denotes the facts of a case and $Q$ denotes the legal question by the user, the \textit{Temporally Applicable-Law Determination (TALD)} task aims to identify a sequence of legal citations $\boldsymbol{y} = [y_1,\dots,y_n]$ where $y_i$ denotes the correct version of statute $i$ that is legally applicable to answering $Q$ based on $F$.
% Note that if $y_i = -1$, it means that statute $i$ is not cited in the legal reasoning process.
 
% Let $\hat{\boldsymbol{y}} = \{\hat{y}_1, \dots, \hat{y}_n\}$ denote the sequence of statute versions predicted by the model in the TALD task.
% The model performance on TALD can be measured by the TALD accuracy, defined by the matching accuracy between the ground truth $\boldsymbol{y}$ and the prediction $\hat{\boldsymbol{y}}$:
% \begin{equation}
% \label{eq:acc}
% ACC(\hat{\boldsymbol{y}}; \boldsymbol{y}) = \frac{\sum_{i \in N} \boldsymbol{1}(y_i = \hat{y}_i \neq -1)}{|\{ i \mid y_i \neq -1 \text{ or } \hat{y}_i \neq -1\}|}.
% \end{equation}
% The objective of TALD is therefore to maximize the TALD accuracy.

We formalize temporal legal reasoning as the
\textbf{Temporally Applicable-Law Determination}
task. Given the facts of a case $F$, a legal question $Q$,
and a set of candidate legal provisions, TALD requires
identifying which provisions are applicable to the case
and determining their governing statutory versions.

Formally, let $\mathcal{N} = \{1, \ldots, n\}$ denote
the set of candidate legal provisions, where each
$i \in \mathcal{N}$ may have multiple statutory versions.
For a query $(F, Q)$, the reference output is represented
as $\mathbf{y} = [y_1, \ldots, y_n]$, where $y_i$
denotes the applicable version of provision $i$.
We set $y_i = -1$ if the provision is not applicable
to the case. The model produces a corresponding
prediction $\hat{\mathbf{y}} = [\hat{y}_1, \ldots,
\hat{y}_n]$.

We define the \textbf{TALD score} as the proportion of correctly matched provision–version pairs among those selected by either the model or the reference:
\begin{equation*}
\operatorname{TALD}(\hat{\mathbf{y}};\mathbf{y})
=
\frac{
\sum_{i \in \mathcal{N}}
\mathbb{1}(\hat{y}_i = y_i \neq -1)
}{
|\{i \mid y_i \neq -1 \lor \hat{y}_i \neq -1\}|
}.
\label{eq:tald}
\end{equation*}

A provision is counted as correct only when the
model selects both the relevant provision and its
applicable statutory version.

\section{Evaluation Setup}
\label{sec:setup}

\subsection{Research Questions}
\label{sec:rq}

We investigate the capability of LLMs on TALD through three research questions (RQs).
\paragraph{RQ1: Temporal version selection.} \textit{How reliably do LLMs select
the applicable statutory version?}
We evaluate TALD performance on cases governed by
earlier and newer laws and examine systematic
patterns in version-selection errors.
\paragraph{RQ2: Sources of failure.} \textit{What type of weakness drives LLM failures on TALD?} 
We investigate whether these failures can be
explained by limitations in legal knowledge or
general reasoning ability, and whether
temporal-law guidance improves performance.
\paragraph{RQ3: Reasoning behavior.} \textit{How does temporal-law
guidance affect LLMs' reasoning behavior?}
We examine changes in token-level predictive
uncertainty at version-related reasoning positions
to investigate how models respond to such guidance.

\subsection{TALD-Bench Dataset Construction}
\label{sec:dataset}

We construct TALD-Bench from Chinese civil judgments involving the transition to the Civil Code in 2021, which consolidated and revised provisions from multiple predecessor statutes. The choice of this testbed is for the availability of the large-scale officially released judgment dataset, and the quantity of statute versional correspondence involved.

Our dataset comprises 51,956 Chinese-language civil
judgments issued between 2022 and 2024 and collected
from China Judgments Online\footnote{https://wenshu.court.gov.cn/}. We extract case facts
and plaintiff claims as model inputs, while the
statutory provisions cited in the judgments serve
as reference labels, allowing us to evaluate model predictions against the legal authorities applied in real judicial decisions. The court's legal reasoning
is excluded from the input to avoid directly
revealing the applicable statutory versions.
Two legally trained coauthors manually checked
approximately 1\% of the processed judgments
for extraction accuracy. Further details are
provided in Appendix~\ref{app:data}.

We divide the dataset into \textit{old-law} and \textit{new-law} splits according to the statutory versions cited in the reference judgments. Old-law cases cite predecessor statutes, such as the Contract Law and Property Law, whereas new-law cases cite corresponding Civil Code provisions. We balance the splits at 25,978 cases each to facilitate comparison across statutory versions. To identify version-selection errors, we construct article-level correspondences between the Civil Code and its predecessor statutes, covering all 1,260 Civil Code articles and their earlier counterparts where available (Appendix~\ref{app:data}).

For RQ2, we additionally construct two diagnostic
knowledge probes: 50 questions assessing the
identification of statutory versions and 16 questions
assessing knowledge of temporal-effect rules (Appendix~\ref{app:knowledge}).

\begin{figure*}[!t]
\centering
\captionsetup[subfloat]{position=top,font={small,sf,bf}}
\subfloat[TALD accuracy.\label{fig:rq1_initial}]{%
\begin{minipage}[t]{0.48\textwidth}
\centering
\begin{tikzpicture}
\definecolor{oldlawbar}{HTML}{245F9E}
\definecolor{newlawbar}{HTML}{A6C8EA}
\begin{axis}[
    tald plot,
    width=0.90\linewidth,
    height=5.0cm,
    ybar=1pt,
    area legend,
    bar width=5pt,
    ymin=0, ymax=1,
    ylabel={TALD accuracy},
    ytick={0,0.2,0.4,0.6,0.8,1},
    yticklabel style={/pgf/number format/fixed},
    xtick={1,2,3,4,5,6,7,8,9},
    xticklabels={
        {DeepSeek-V3.2}, {GLM-4.7}, {GPT-5.4},
        {Gemini-3.1-Pro}, {Claude Opus 4.6},
        {Qwen3-235B}, {Qwen3-80B}, {Qwen3-30B},
        {LegalOne-8B}
    },
    xticklabel style={align=right, rotate=45, anchor=east, font=\scriptsize\sffamily},
    tick label style={font=\small\sffamily},
    label style={font=\small\sffamily},
    enlarge x limits=0.065,
    axis lines*=left,
    axis line style={taldaxis, line width=0.6pt},
    tick style={taldaxis, line width=0.6pt},
    ymajorgrids=true,
    grid style={taldgrid, line width=0.5pt},
    legend style={
        at={(0.5,1.03)}, anchor=south, legend columns=1,
        draw=none, font=\scriptsize\sffamily, /tikz/every even column/.append style={column sep=3pt}
    },
    nodes near coords,
    every node near coord/.append style={
        font=\fontsize{6}{7}\selectfont\sffamily, text=black,
        /pgf/number format/fixed,
        /pgf/number format/precision=3,
        /pgf/number format/fixed zerofill
    }
]
\addplot[fill=oldlawbar, draw=oldlawbar!70!black, line width=1.4pt] coordinates {
    (1,0.092) (2,0.080) (3,0.237) (4,0.135) (5,0.014)
    (6,0.148) (7,0.028) (8,0.020) (9,0.093)
};
\addplot[fill=newlawbar, draw=newlawbar!70!black, line width=1.4pt] coordinates {
    (1,0.746) (2,0.787) (3,0.701) (4,0.843) (5,0.820)
    (6,0.730) (7,0.798) (8,0.765) (9,0.800)
};
\legend{old-law cases, new-law cases}
\end{axis}
\end{tikzpicture}
\end{minipage}}\hfill
\subfloat[Weighted error distribution.\label{fig:fault_types}]{%
\begin{minipage}[t]{0.48\textwidth}
\centering
\begin{tikzpicture}
\definecolor{oldnewbar}{HTML}{245F9E}
\definecolor{newoldbar}{HTML}{A6C8EA}
\definecolor{irrelevantbar}{HTML}{897675}
\begin{axis}[
    tald plot,
    width=0.90\linewidth,
    height=5.0cm,
    ybar stacked,
    area legend,
    bar width=11pt,
    ymin=0, ymax=100,
    ylabel={Share of weighted errors (\%)},
    ytick={0,20,40,60,80,100},
    xtick={1,2,3,4,5,6,7,8,9},
    xticklabels={
        {DeepSeek-V3.2}, {GLM-4.7}, {GPT-5.4},
        {Gemini-3.1-Pro}, {Claude Opus 4.6},
        {Qwen3-235B}, {Qwen3-80B}, {Qwen3-30B},
        {LegalOne-8B}
    },
    xticklabel style={align=right, rotate=45, anchor=east, font=\scriptsize\sffamily},
    tick label style={font=\small\sffamily},
    label style={font=\small\sffamily},
    enlarge x limits=0.065,
    axis lines*=left,
    axis line style={taldaxis, line width=0.6pt},
    tick style={taldaxis, line width=0.6pt},
    ymajorgrids=true,
    grid style={taldgrid, line width=0.5pt},
    legend style={
        at={(0.5,1.03)}, anchor=south, legend columns=2,
        draw=none, font=\scriptsize\sffamily, /tikz/every even column/.append style={column sep=3pt}
    }
]
\addplot[fill=oldnewbar, draw=oldnewbar!70!black, line width=1.4pt] coordinates {
    (1,85) (2,89) (3,69) (4,79) (5,89)
    (6,75) (7,87) (8,79) (9,94)
};
\addplot[fill=newoldbar, draw=newoldbar!70!black, line width=1.4pt] coordinates {
    (1,3) (2,0) (3,14) (4,6) (5,0)
    (6,5) (7,0) (8,2) (9,0)
};
\addplot[fill=irrelevantbar, draw=irrelevantbar!70!black, line width=1.4pt] coordinates {
    (1,12) (2,11) (3,17) (4,15) (5,11)
    (6,20) (7,13) (8,19) (9,6)
};
\legend{old-to-new error, new-to-old error, irrelevant-law error}
\end{axis}
\end{tikzpicture}
\end{minipage}}
\caption{TALD accuracy on old-law and new-law cases (Equation~\ref{eq:tald}) and the distribution of weighted errors pooled across both splits. Each stacked bar sums to 100\%.}
\label{fig:rq1_results}
\end{figure*}

\subsection{General Experimental Setups}
\label{sec:expsetup}

\myparagraph{Models.}
We evaluate nine LLMs, covering both edge-cutting commercial
and open-weight models: GPT-5.4~\cite{openai2026gpt5systemcard}, Claude Opus 4.6~\cite{anthropic2026claudeopus46},
Gemini-3.1-Pro~\cite{deepmind2026gemini31pro}, DeepSeek-V3.2~\cite{deepseek2025v32}, GLM-4.7~\cite{zai2025glm47},
Qwen3-235B, Qwen3-80B, Qwen3-30B~\cite{qwen2025qwen3}, and
LegalOne-8B~\cite{li2026legalonefamilyfoundationmodels}. We evaluate reasoning
and non-reasoning configurations where applicable.
Detailed model identifiers and configurations
are provided in Appendix~\ref{app:models}.

\myparagraph{Evaluation Protocol.}
For the primary TALD experiments in RQ1 and RQ2,
we randomly sample 100 cases from each split per
run and repeat the evaluation with four sampling
seeds. Given the substantial inference costs of evaluating nine LLMs across multiple reasoning configurations and prompting conditions, we use sampled evaluations rather than exhaustively evaluating the full benchmark. The same sampled cases are shared across
models and prompt conditions within each run,
ensuring paired comparisons.
For RQ3, we evaluate 100 old-law cases per run
across three sampling seeds using locally deployed
models to obtain token-level probability distributions.
We report mean TALD scores across runs, with
standard errors calculated over sampling seeds.
Generation parameters and implementation details
are provided in Appendix~\ref{app:runs}.

% TODO: Report the actual number of distinct cases
% evaluated in each split.
% TODO: Explain why 100 cases per split were sampled.
% TODO: Consider a larger-sample robustness evaluation.

% TODO: Implement and report baseline results in RQ1.
% TODO: Specify how candidates without a Civil Code
% counterpart and irrelevant candidates are handled.

% Queue the knowledge probes for the following RQ2 discussion.

\section{Evaluation Results}
\label{sec:results}
\subsection{Results on Temporal Version Selection}
\label{sec:rq1}
\label{ssec:rq1_results}

\myparagraph{Experiment 1: Temporal version selection by split.}
RQ1 asks \textit{whether} LLMs can determine the applicable statutory version from a case's temporal information. We evaluate models in their competitive thinking mode or high reasoning effort setting. Because reliable TALD requires selecting the governing version whether earlier or newer law applies, we report accuracy separately on the old-law and new-law splits and analyze the direction of version-selection errors.

\begin{figure*}[!t]
\centering
\captionsetup[subfloat]{position=top,font={small,sf,bf}}
\subfloat[Statutory-version knowledge.\label{fig:statutory_version_knowledge}]{%
\begin{minipage}[t]{0.44\textwidth}
\centering
\begin{tikzpicture}
\definecolor{knowledgebar}{HTML}{245F9E}
\begin{axis}[
    tald plot,
    width=\linewidth,
    height=4.5cm,
    ybar,
    bar width=10pt,
    ymin=0, ymax=1.12,
    ylabel={Accuracy},
    ytick={0,0.2,0.4,0.6,0.8,1},
    yticklabel style={/pgf/number format/fixed},
    xtick={1,2,3,4,5,6,7,8,9},
    xticklabels={
        {DeepSeek-V3.2}, {GLM-4.7}, {GPT-5.4},
        {Gemini-3.1-Pro}, {Claude Opus 4.6},
        {Qwen3-235B}, {Qwen3-80B}, {Qwen3-30B}, {LegalOne-8B}
    },
    xticklabel style={rotate=45, anchor=east, font=\scriptsize\sffamily},
    tick label style={font=\small\sffamily},
    label style={font=\small\sffamily},
    enlarge x limits=0.065,
    axis lines*=left,
    axis line style={taldaxis, line width=0.6pt},
    tick style={taldaxis, line width=0.6pt},
    ymajorgrids=true,
    grid style={taldgrid, line width=0.5pt},
    nodes near coords,
    every node near coord/.append style={
        font=\scriptsize\sffamily, text=black,
        /pgf/number format/fixed,
        /pgf/number format/precision=2,
        /pgf/number format/fixed zerofill
    }
]
\addplot[fill=knowledgebar, draw=knowledgebar!70!black, line width=1.4pt] coordinates {
    (1,0.88) (2,0.98) (3,0.98) (4,1.00) (5,0.92) (6,0.98) (7,1.00) (8,0.90) (9,1.00)
};
\end{axis}
\end{tikzpicture}
\end{minipage}}\hfill
\subfloat[Temporal-effect rule knowledge.\label{fig:temporal_effect_knowledge}]{%
\begin{minipage}[t]{0.44\textwidth}
\centering
\begin{tikzpicture}
\definecolor{knowledgebar}{HTML}{A6C8EA}
\begin{axis}[
    tald plot,
    width=\linewidth,
    height=4.5cm,
    ybar,
    bar width=10pt,
    ymin=0, ymax=1.12,
    ylabel={Accuracy},
    ytick={0,0.2,0.4,0.6,0.8,1},
    yticklabel style={/pgf/number format/fixed},
    xtick={1,2,3,4,5,6,7,8,9},
    xticklabels={
        {DeepSeek-V3.2}, {GLM-4.7}, {GPT-5.4},
        {Gemini-3.1-Pro}, {Claude Opus 4.6},
        {Qwen3-235B}, {Qwen3-80B}, {Qwen3-30B}, {LegalOne-8B}
    },
    xticklabel style={rotate=45, anchor=east, font=\scriptsize\sffamily},
    tick label style={font=\small\sffamily},
    label style={font=\small\sffamily},
    enlarge x limits=0.065,
    axis lines*=left,
    axis line style={taldaxis, line width=0.6pt},
    tick style={taldaxis, line width=0.6pt},
    ymajorgrids=true,
    grid style={taldgrid, line width=0.5pt},
    nodes near coords,
    every node near coord/.append style={
        font=\scriptsize\sffamily, text=black,
        /pgf/number format/fixed,
        /pgf/number format/precision=2,
        /pgf/number format/fixed zerofill
    }
]
% Fixed 16-item probe: Gemini-3.1-Pro scored 15/16 = 0.9375 (shown as 0.94).
\addplot[fill=knowledgebar, draw=knowledgebar!70!black, line width=1.4pt] coordinates {
    (1,0.94) (2,1.00) (3,0.94) (4,0.94) (5,0.94) (6,1.00) (7,0.81) (8,0.88) (9,0.88)
};
\end{axis}
\end{tikzpicture}
\end{minipage}}
\caption{Accuracy on the two legal knowledge probes.}
\label{tab:knowledge_probes}
\end{figure*}
% Figure artwork generated by others/plot_hint_correlation_heatmaps.py.
\begin{figure*}[t]
\centering
\includegraphics[width=0.85\textwidth]{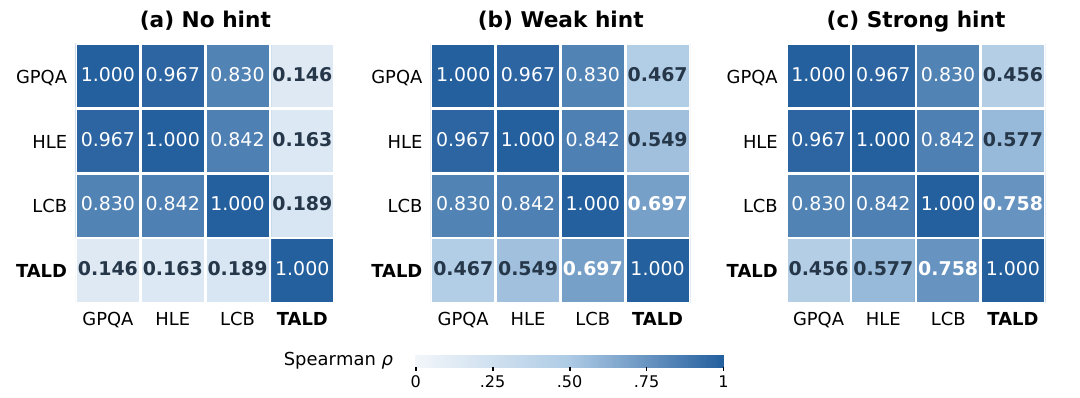}
% FIG4-REVIEW F01 [applied]: Three panels replace the heatmap/line-chart pair.
% Each panel uses the same color scale; self-correlations are now shown as 1.
\caption{Spearman correlations among GPQA, HLE, LCB, and old-law TALD accuracy under (a) no hint, (b) weak hint, and (c) strong hint. All panels share a color scale; diagonal entries display self-correlations (1.000). External benchmark scores and matched configurations are fixed across hint conditions. Details see Appendix~\ref{app:reasoning_benchmarks}.}
\label{fig:reasoning_correlations}
\end{figure*}

\myparagraph{Finding 1: LLMs achieve much higher TALD accuracy on new-law cases than on old-law cases (Figure~\ref{fig:rq1_initial}; full results in Appendix~\ref{app:additional_r1}).}
\label{finding:1}
All nine models exceed $0.70$ accuracy on the new-law split but fall below $0.25$ on the old-law split. Strong performance on new-law cases thus masks frequent failures when earlier law governs.

\myparagraph{Experiment 2: Error types in version selection.}
This gap suggests that models may systematically favor newer law, which helps on new-law cases but hurts on old-law cases. We test this hypothesis with an \textit{Error-Type Analysis} that classifies each case in which a model misses the required law version: an \textsc{Old-to-New} error misses a required earlier law and cites the Civil Code instead; a \textsc{New-to-Old} error misses a required Civil Code version and cites an earlier statute instead; and an \textsc{Irrelevant-Law} error misses the required law without citing any version of it. For each model, we pool errors across both splits after normalizing each case's error counts by its number of required laws (Appendix~\ref{app:fault_types}).

\myparagraph{Finding 2: Most TALD errors come from applying newer law to old-law cases (Figure~\ref{fig:fault_types}).}
\textsc{Old-to-New} errors, which apply the Civil Code to cases governed by earlier law, account for 69--94\% of weighted errors across models, whereas \textsc{New-to-Old} errors account for at most 14\%. The performance gap in Finding~1 therefore stems mainly from a systematic preference for newer law.

\subsection{Results on Sources of TALD Failure}
\label{sec:rq2}

% Two-column floats need to be queued ahead of their discussion.
%\input{figures/reasoning_correlations.tex}

RQ1 reveals failures on old-law cases and a systematic preference for newer law. This motivates RQ2: \textit{what weakness} limits their ability to determine the temporally applicable law? We consider three capabilities that LLMs acquire during training: knowledge, instruction following, and reasoning~\citep{petroni2019language,zhou2023instruction,suzgun2023challenging}. Because models return the requested legal provisions and perform relatively well on new-law cases, basic instruction following is unlikely to explain the errors. We therefore test whether LLMs possess the required legal knowledge and how well their reasoning capabilities supports TALD.

\myparagraph{Experiment 3: Weakness probe on legal knowledge.}
\label{ssec:rq2_knowledge}
Given that \textsc{Old-to-New} errors dominate TALD failures, a natural hypothesis is that poor performance on old-law cases
stems from insufficient knowledge of earlier provisions, confusion between versions, or a lack of knowledge about temporal applicability. If so, models should struggle to distinguish revised provisions or to identify when each version applies. We test both abilities with two multiple-choice question (MCQ) probes (See Appendices~\ref{app:memorization} and~\ref{app:mcq} for details).

The \textit{statutory-version probe} draws on correspondences between the Civil Code and its predecessor statutes. Our legal experts select provisions with major substantive revisions, so each pair contrasts two rules on the same legal issue. Each of the 50 questions presents one provision and asks whether it belongs to the Civil Code or the corresponding earlier statute.

The \textit{temporal-effect rule probe} contains 16 questions based on the Supreme People's Court's judicial interpretation of the Civil Code's temporal effect~\citep{spc2020temporal}. Each asks for the legally correct option under a specified temporal-effect condition.

\begin{table*}[t]
\centering
\begingroup
% Two balanced panels: 11 configurations on each side.
\fontsize{8}{10}\selectfont
\setlength{\tabcolsep}{1.5pt}
\renewcommand{\arraystretch}{1.0}
% Report the mean and across-seed standard error to three decimal places.
\newcommand{\taldresult}[2]{#1\,{\fontsize{6.5}{7.5}\selectfont$\pm$\,#2}}
\begin{minipage}[t]{0.485\textwidth}
\vspace{0pt}
\begin{tabular*}{\linewidth}{@{\extracolsep{\fill}}llccc@{}}
\toprule
\textbf{Model} & \textbf{Mode} & \textbf{Default} & \textbf{Weak} & \textbf{Strong} \\
\midrule
\multirow{4}{*}{Claude Opus 4.6} & low & \taldresult{0.021}{0.001} & \taldresult{0.131}{0.011} & \taldresult{0.185}{0.107} \\
 & medium & \taldresult{0.019}{0.002} & \taldresult{0.102}{0.008} & \taldresult{0.143}{0.083} \\
 & high & \taldresult{0.026}{0.002} & \taldresult{0.099}{0.024} & \taldresult{0.154}{0.089} \\
 & max & \taldresult{0.014}{0.004} & \taldresult{0.113}{0.033} & \taldresult{0.156}{0.091} \\
\addlinespace[6.25pt]
\multirow{3}{*}{Gemini-3.1-Pro} & low & \taldresult{0.146}{0.011} & \taldresult{0.501}{0.019} & \taldresult{0.439}{0.035} \\
 & medium & \taldresult{0.131}{0.020} & \taldresult{0.503}{0.021} & \taldresult{0.489}{0.016} \\
 & high & \taldresult{0.135}{0.013} & \taldresult{0.474}{0.026} & \taldresult{0.486}{0.010} \\
\addlinespace[6.25pt]
\multirow{4}{*}{GPT-5.4} & none & \taldresult{0.021}{0.005} & \taldresult{0.087}{0.013} & \taldresult{0.167}{0.021} \\
 & low & \taldresult{0.082}{0.030} & \taldresult{0.444}{0.032} & \taldresult{0.456}{0.016} \\
 & medium & \taldresult{0.196}{0.032} & \taldresult{0.472}{0.002} & \taldresult{0.486}{0.008} \\
 & high & \taldresult{0.237}{0.005} & \taldresult{0.416}{0.000} & \taldresult{0.457}{0.003} \\
\bottomrule
\end{tabular*}
\end{minipage}\hfill
\begin{minipage}[t]{0.485\textwidth}
\vspace{0pt}
\begin{tabular*}{\linewidth}{@{\extracolsep{\fill}}llccc@{}}
\toprule
\textbf{Model} & \textbf{Mode} & \textbf{Default} & \textbf{Weak} & \textbf{Strong} \\
\midrule
\multirow{2}{*}{DeepSeek-V3.2} & chat & \taldresult{0.013}{0.003} & \taldresult{0.028}{0.005} & \taldresult{0.080}{0.010} \\
 & reasoner & \taldresult{0.092}{0.017} & \taldresult{0.394}{0.013} & \taldresult{0.436}{0.022} \\
\addlinespace[2.5pt]
\multirow{2}{*}{GLM-4.7} & instruct & \taldresult{0.013}{0.007} & \taldresult{0.032}{0.005} & \taldresult{0.053}{0.008} \\
 & thinking & \taldresult{0.080}{0.024} & \taldresult{0.337}{0.019} & \taldresult{0.435}{0.027} \\
\addlinespace[2.5pt]
\multirow{2}{*}{Qwen3-235B} & instruct & \taldresult{0.035}{0.005} & \taldresult{0.035}{0.008} & \taldresult{0.041}{0.009} \\
 & thinking & \taldresult{0.148}{0.017} & \taldresult{0.330}{0.010} & \taldresult{0.323}{0.010} \\
\addlinespace[2.5pt]
\multirow{2}{*}{Qwen3-80B} & instruct & \taldresult{0.080}{0.006} & \taldresult{0.078}{0.001} & \taldresult{0.066}{0.004} \\
 & thinking & \taldresult{0.028}{0.005} & \taldresult{0.386}{0.006} & \taldresult{0.512}{0.016} \\
\addlinespace[2.5pt]
\multirow{2}{*}{Qwen3-30B} & instruct & \taldresult{0.292}{0.008} & \taldresult{0.264}{0.024} & \taldresult{0.164}{0.013} \\
 & thinking & \taldresult{0.020}{0.008} & \taldresult{0.350}{0.009} & \taldresult{0.395}{0.015} \\
\addlinespace[2.5pt]
LegalOne-8B & reasoning & \taldresult{0.093}{0.008} & \taldresult{0.280}{0.033} & \taldresult{0.298}{0.011} \\
\bottomrule
\end{tabular*}
\end{minipage}
\endgroup
\caption{Effect of temporal-law hints on old-law cases, measured by TALD accuracy. Each entry reports the mean $\pm$ its across-seed standard error. Default denotes the no-hint setting.}
\label{tab:default_reasoning}
\end{table*}

\myparagraph{Finding 3: Models perform well on isolated legal knowledge probes despite poor TALD performance (Figures~\ref{fig:statutory_version_knowledge} and~\ref{fig:temporal_effect_knowledge}).}
Models achieve 0.88--1.00 accuracy on the statutory-version probe (Figure~\ref{fig:statutory_version_knowledge}), far above their accuracy on old-law cases: they usually recognize a provision's statutory version but struggle to select the applicable version from case facts. Likewise, all models exceed 0.80 accuracy on the temporal-effect rule probe, and several reach about 0.90 or higher (Figure~\ref{fig:temporal_effect_knowledge}); models that often fail on old-law cases still answer many abstract temporal-effect questions correctly. Together, the two probes suggest that missing legal knowledge is unlikely to be the principal cause of TALD failure.

% Queue the hint-comparison table before Findings 4 and 5.

\myparagraph{Experiment 4: Weakness probe on general reasoning ability.}
\label{ssec:rq2_reasoning}
\label{sssec:rq2_general_reasoning}
TALD demands substantial reasoning. For example, models must map case details onto abstract temporal-effect rules and compare the time spans of legal facts with the validity periods of different statutory versions.
A model may therefore have the relevant legal knowledge yet lack the general reasoning ability to apply it to complex cases. If so, stronger reasoning should predict better TALD performance, both across models and across configurations within a model family.

We compare LLMs' TALD accuracy with their publicly reported scores on three widely used reasoning benchmarks, GPQA, HLE, and LiveCodeBench, which measure scientific, academic, and coding reasoning, respectively~\citep{artificialanalysis2026methodology}. We first evaluate the target LLMs on TALD under different
reasoning modes (the ``Default'' column of Table~\ref{tab:default_reasoning}). We then match scores by model and reasoning configuration and compute Spearman's $\rho$. Appendix~\ref{app:reasoning_benchmarks} details the settings.

\begin{figure*}[!t]
\centering
\captionsetup[subfloat]{position=top,font={scriptsize,sf,bf}}
\definecolor{rqthreeacc}{HTML}{5B7B99}
\definecolor{rqthreeentropy}{HTML}{B92E26}
\subfloat[Qwen3-235B (thinking).\label{fig:rq3_qwen235}]{%
\begin{minipage}[t]{0.24\textwidth}
\centering
\begin{tikzpicture}[baseline={(0,0)}]
% Equal-width panel frame; center the 0.54-linewidth plot within it.
\path[use as bounding box] (0,-0.8cm) rectangle (\linewidth,3.1cm);
\begin{axis}[
    tald plot, at={(0.23\linewidth,0)}, anchor=south west,
    width=0.54\linewidth, height=2.9cm, scale only axis,
    xmin=-0.20, xmax=2.20,
    xtick={0,1,2}, xticklabels={No hint,Weak hint,Strong hint}, xticklabel style={font=\scriptsize\sffamily,text=black,rotate=25,anchor=east},
    tick label style={font=\scriptsize\sffamily}, label style={font=\scriptsize\sffamily},
    axis line style={taldaxis, line width=0.6pt}, tick style={taldaxis, line width=0.6pt}, tick align=outside, clip=false,
    axis y line*=left, axis x line*=bottom,
    ymin=0, ymax=0.68, ytick={0,0.2,0.4,0.6},
    yticklabel style={text=rqthreeacc},
    ymajorgrids=true, grid style={taldgrid, line width=0.5pt},
    legend to name=rqthreelegend, legend columns=2,
    legend style={draw=none,font=\scriptsize\sffamily,/tikz/every even column/.append style={column sep=1.5em}},
]
\addplot[color=rqthreeacc, line width=1.2pt, mark=*, mark size=2.2pt, error bars/.cd, y dir=both, y explicit] coordinates {(0,0.155) +- (0,0.025) (1,0.524) +- (0,0.025) (2,0.523) +- (0,0.032)};
\addlegendentry{TALD accuracy}
\addlegendimage{color=rqthreeentropy, line width=1.2pt, mark=square*}
\addlegendentry{Entropy}
\end{axis}
\begin{axis}[
    tald plot, at={(0.23\linewidth,0)}, anchor=south west,
    width=0.54\linewidth, height=2.9cm, scale only axis,
    xmin=-0.20, xmax=2.20,
    xtick={0,1,2}, xticklabels={No hint,Weak hint,Strong hint}, xticklabel style={font=\scriptsize\sffamily,text=black,rotate=25,anchor=east},
    tick label style={font=\scriptsize\sffamily}, label style={font=\scriptsize\sffamily},
    axis line style={taldaxis, line width=0.6pt}, tick style={taldaxis, line width=0.6pt}, tick align=outside, clip=false,
    axis y line*=right, axis x line=none, ymajorgrids=false,
    ymin=0.22, ymax=0.28, ytick={0.22,0.24,0.26,0.28},
    yticklabel style={text=rqthreeentropy, /pgf/number format/fixed,
        /pgf/number format/precision=2, /pgf/number format/fixed zerofill},
]
\addplot[color=rqthreeentropy, line width=1.2pt, mark=square*, mark size=2pt, error bars/.cd, y dir=both, y explicit] coordinates {(0,0.238) +- (0,0.006) (1,0.253) +- (0,0.004) (2,0.266) +- (0,0.005)};
\end{axis}
\end{tikzpicture}
\end{minipage}}
\hfill%
\subfloat[Qwen3-80B (thinking).\label{fig:rq3_qwen80}]{%
\begin{minipage}[t]{0.24\textwidth}
\centering
\begin{tikzpicture}[baseline={(0,0)}]
% Equal-width panel frame; center the 0.54-linewidth plot within it.
\path[use as bounding box] (0,-0.8cm) rectangle (\linewidth,3.1cm);
\begin{axis}[
    tald plot, at={(0.23\linewidth,0)}, anchor=south west,
    width=0.54\linewidth, height=2.9cm, scale only axis,
    xmin=-0.20, xmax=2.20,
    xtick={0,1,2}, xticklabels={No hint,Weak hint,Strong hint}, xticklabel style={font=\scriptsize\sffamily,text=black,rotate=25,anchor=east},
    tick label style={font=\scriptsize\sffamily}, label style={font=\scriptsize\sffamily},
    axis line style={taldaxis, line width=0.6pt}, tick style={taldaxis, line width=0.6pt}, tick align=outside, clip=false,
    axis y line*=left, axis x line*=bottom,
    ymin=0, ymax=0.68, ytick={0,0.2,0.4,0.6},
    yticklabel style={text=rqthreeacc},
    ymajorgrids=true, grid style={taldgrid, line width=0.5pt},
]
\addplot[color=rqthreeacc, line width=1.2pt, mark=*, mark size=2.2pt, error bars/.cd, y dir=both, y explicit] coordinates {(0,0.050) +- (0,0.002) (1,0.459) +- (0,0.022) (2,0.578) +- (0,0.024)};
\end{axis}
\begin{axis}[
    tald plot, at={(0.23\linewidth,0)}, anchor=south west,
    width=0.54\linewidth, height=2.9cm, scale only axis,
    xmin=-0.20, xmax=2.20,
    xtick={0,1,2}, xticklabels={No hint,Weak hint,Strong hint}, xticklabel style={font=\scriptsize\sffamily,text=black,rotate=25,anchor=east},
    tick label style={font=\scriptsize\sffamily}, label style={font=\scriptsize\sffamily},
    axis line style={taldaxis, line width=0.6pt}, tick style={taldaxis, line width=0.6pt}, tick align=outside, clip=false,
    axis y line*=right, axis x line=none, ymajorgrids=false,
    ymin=1.1, ymax=1.26, ytick={1.10,1.15,1.20,1.25},
    yticklabel style={text=rqthreeentropy, /pgf/number format/fixed,
        /pgf/number format/precision=2, /pgf/number format/fixed zerofill},
]
\addplot[color=rqthreeentropy, line width=1.2pt, mark=square*, mark size=2pt, error bars/.cd, y dir=both, y explicit] coordinates {(0,1.148) +- (0,0.026) (1,1.185) +- (0,0.029) (2,1.211) +- (0,0.025)};
\end{axis}
\end{tikzpicture}
\end{minipage}}
\hfill%
\subfloat[Qwen3-30B (thinking).\label{fig:rq3_qwen30}]{%
\begin{minipage}[t]{0.24\textwidth}
\centering
\begin{tikzpicture}[baseline={(0,0)}]
% Equal-width panel frame; center the 0.54-linewidth plot within it.
\path[use as bounding box] (0,-0.8cm) rectangle (\linewidth,3.1cm);
\begin{axis}[
    tald plot, at={(0.23\linewidth,0)}, anchor=south west,
    width=0.54\linewidth, height=2.9cm, scale only axis,
    xmin=-0.20, xmax=2.20,
    xtick={0,1,2}, xticklabels={No hint,Weak hint,Strong hint}, xticklabel style={font=\scriptsize\sffamily,text=black,rotate=25,anchor=east},
    tick label style={font=\scriptsize\sffamily}, label style={font=\scriptsize\sffamily},
    axis line style={taldaxis, line width=0.6pt}, tick style={taldaxis, line width=0.6pt}, tick align=outside, clip=false,
    axis y line*=left, axis x line*=bottom,
    ymin=0, ymax=0.68, ytick={0,0.2,0.4,0.6},
    yticklabel style={text=rqthreeacc},
    ymajorgrids=true, grid style={taldgrid, line width=0.5pt},
]
\addplot[color=rqthreeacc, line width=1.2pt, mark=*, mark size=2.2pt, error bars/.cd, y dir=both, y explicit] coordinates {(0,0.020) +- (0,0.015) (1,0.371) +- (0,0.015) (2,0.450) +- (0,0.020)};
\end{axis}
\begin{axis}[
    tald plot, at={(0.23\linewidth,0)}, anchor=south west,
    width=0.54\linewidth, height=2.9cm, scale only axis,
    xmin=-0.20, xmax=2.20,
    xtick={0,1,2}, xticklabels={No hint,Weak hint,Strong hint}, xticklabel style={font=\scriptsize\sffamily,text=black,rotate=25,anchor=east},
    tick label style={font=\scriptsize\sffamily}, label style={font=\scriptsize\sffamily},
    axis line style={taldaxis, line width=0.6pt}, tick style={taldaxis, line width=0.6pt}, tick align=outside, clip=false,
    axis y line*=right, axis x line=none, ymajorgrids=false,
    ymin=0.29, ymax=0.5, ytick={0.30,0.35,0.40,0.45,0.50},
    yticklabel style={text=rqthreeentropy, /pgf/number format/fixed,
        /pgf/number format/precision=2, /pgf/number format/fixed zerofill},
]
\addplot[color=rqthreeentropy, line width=1.2pt, mark=square*, mark size=2pt, error bars/.cd, y dir=both, y explicit] coordinates {(0,0.315) +- (0,0.005) (1,0.425) +- (0,0.010) (2,0.462) +- (0,0.014)};
\end{axis}
\end{tikzpicture}
\end{minipage}}
\hfill%
\subfloat[LegalOne-8B (reasoning).\label{fig:rq3_legalone}]{%
\begin{minipage}[t]{0.24\textwidth}
\centering
\begin{tikzpicture}[baseline={(0,0)}]
% Equal-width panel frame; center the 0.54-linewidth plot within it.
\path[use as bounding box] (0,-0.8cm) rectangle (\linewidth,3.1cm);
\begin{axis}[
    tald plot, at={(0.23\linewidth,0)}, anchor=south west,
    width=0.54\linewidth, height=2.9cm, scale only axis,
    xmin=-0.20, xmax=2.20,
    xtick={0,1,2}, xticklabels={No hint,Weak hint,Strong hint}, xticklabel style={font=\scriptsize\sffamily,text=black,rotate=25,anchor=east},
    tick label style={font=\scriptsize\sffamily}, label style={font=\scriptsize\sffamily},
    axis line style={taldaxis, line width=0.6pt}, tick style={taldaxis, line width=0.6pt}, tick align=outside, clip=false,
    axis y line*=left, axis x line*=bottom,
    ymin=0, ymax=0.68, ytick={0,0.2,0.4,0.6},
    yticklabel style={text=rqthreeacc},
    ymajorgrids=true, grid style={taldgrid, line width=0.5pt},
]
\addplot[color=rqthreeacc, line width=1.2pt, mark=*, mark size=2.2pt, error bars/.cd, y dir=both, y explicit] coordinates {(0,0.093) +- (0,0.008) (1,0.280) +- (0,0.032) (2,0.298) +- (0,0.011)};
\end{axis}
\begin{axis}[
    tald plot, at={(0.23\linewidth,0)}, anchor=south west,
    width=0.54\linewidth, height=2.9cm, scale only axis,
    xmin=-0.20, xmax=2.20,
    xtick={0,1,2}, xticklabels={No hint,Weak hint,Strong hint}, xticklabel style={font=\scriptsize\sffamily,text=black,rotate=25,anchor=east},
    tick label style={font=\scriptsize\sffamily}, label style={font=\scriptsize\sffamily},
    axis line style={taldaxis, line width=0.6pt}, tick style={taldaxis, line width=0.6pt}, tick align=outside, clip=false,
    axis y line*=right, axis x line=none, ymajorgrids=false,
    ymin=0.135, ymax=0.185, ytick={0.14,0.15,0.16,0.17,0.18},
    yticklabel style={text=rqthreeentropy, /pgf/number format/fixed,
        /pgf/number format/precision=2, /pgf/number format/fixed zerofill},
]
\addplot[color=rqthreeentropy, line width=1.2pt, mark=square*, mark size=2pt, error bars/.cd, y dir=both, y explicit] coordinates {(0,0.147) +- (0,0.006) (1,0.159) +- (0,0.002) (2,0.173) +- (0,0.003)};
\end{axis}
\end{tikzpicture}
\end{minipage}}
\par\vspace{0.2em}
\pgfplotslegendfromname{rqthreelegend}
\caption{TALD accuracy and version-related policy entropy on old-law cases across hint strength, shown separately for each model. Points show means over three random seeds; error bars show standard errors. TALD accuracy uses a common left-axis range; entropy uses model-specific right-axis ranges. Full results are reported in Table~\ref{tab:app_rq3_main}.}
\label{fig:rq3_main}
\end{figure*}

\myparagraph{Finding 4a: Without hint intervention, general reasoning performance shows little correlation with TALD performance (Figure~\ref{fig:reasoning_correlations}a).}
%Across matched configurations, GPQA and HLE correlate weakly with default TALD accuracy ($\rho=0.146$ and $0.163$) but strongly with each other ($\rho=0.967$; Figure~\ref{fig:reasoning_correlations}a). LiveCodeBench also correlates strongly with GPQA and HLE ($\rho=0.830$ and $0.842$) but weakly with TALD ($\rho=0.189$). Thus, general reasoning capabilities are not well aligned with TALD performance; the improvement of the former is faintly related to the enhancement of the latter. 
The no-hint correlation heatmap reveals a broad mismatch between general reasoning performance and TALD. GPQA, HLE, and LiveCodeBench correlate strongly with one another ($\rho=0.830$--$0.967$) but weakly with default TALD accuracy ($\rho=0.146$, $0.163$, and $0.189$, respectively). Models that rank high on general reasoning benchmarks are thus not necessarily those that rank high on TALD.

\myparagraph{Finding 4b: Increasing reasoning effort rarely improves TALD performance (Table~\ref{tab:default_reasoning}).}
Within-family comparisons under the default setting point the same way. Raising the reasoning effort yields no consistent improvement for Claude Opus 4.6, whose accuracy fluctuates between $0.014$ and $0.026$ from low to max effort, or for Gemini-3.1-Pro, which stays between $0.131$ and $0.146$ from low to high effort. Enabling the thinking mode even lowers accuracy for Qwen3-30B ($0.292 \to 0.020$) and Qwen3-80B ($0.080 \to 0.028$). Where more reasoning does help, as for GPT-5.4 (from $0.021$ without reasoning to $0.237$ at high effort) and the thinking variants of DeepSeek-V3.2, GLM-4.7, and Qwen3-235B, accuracy remains below $0.24$.
%Qwen3-30B Instruct achieves the highest old-law accuracy among default configurations ($0.292$), despite failing much behind on general reasoning benchmarks compared to large reasoning LLMs.Within LLM families, as is observed in Table~\ref{tab:default_reasoning} and Table~\ref{tab:app_external_reasoning}, although (more) thinking generally improves general reasoning performances,TALD accuracy is observed to reduce for Qwen3-30B and Qwen3-80B when the thinking mode is turned on; similarly, higher effort Claude and Gemini show no consistent improvement. 
Notably, the highest default old-law accuracy comes from Qwen3-30B Instruct, a relatively small non-thinking model ($0.292$), which exceeds Qwen3-235B thinking ($0.148$), GPT-5.4 high ($0.237$), and Claude Opus 4.6 max ($0.014$). This result is striking because Qwen3-30B Instruct trails many of these models on the matched reasoning benchmarks: the strongest general reasoners fail to outperform it on TALD.

Although some configurations benefit from more reasoning, the overall mismatch raises a central question: why do strong reasoning models lose their expected advantage on TALD?

\myparagraph{Experiment 5: Weakness probe on domain-specific reasoning policy.}
\label{sssec:rq2_hint}
One possible obstacle is the model's default \emph{domain-specific reasoning policy}: how it uses case facts and legal rules to select a statutory version. The newer-law bias found in RQ1 suggests that this policy may favor familiar newer law without adequately checking its temporal applicability, which would limit even a strong reasoner. If so, directing attention to temporal applicability should improve TALD and reveal reasoning advantages hidden under the default instruction.

Our legal experts construct weak and strong temporal-law hints based on laws, judicial interpretations, and expertise. The weak hint directs attention to event timing and retroactivity; the strong hint adds a brief account of non-retroactivity, its exceptions, and the use of newer provisions in judicial reasoning (Appendix~\ref{app:prompts}). We compare both hints with the default instruction on the same old-law cases, keeping all other settings fixed. The weak hint is especially informative: it gives neither a case-specific answer nor a rule for choosing a version, so it tests whether a simple reminder can elicit models' existing reasoning ability.

\myparagraph{Finding 5: Temporal-law hints strengthen the association between general reasoning performance and TALD accuracy (Figure~\ref{fig:reasoning_correlations}).}
The correlations between TALD and general reasoning scores are notably stronger in both hint heatmaps than in the no-hint heatmap. Under the weak and strong hints, Spearman correlations rise to $0.467$ and $0.456$ for GPQA, $0.549$ and $0.577$ for HLE, and $0.697$ and $0.758$ for LiveCodeBench(LCB), respectively. Both hint conditions exceed the no-hint correlations on all three benchmarks, although the GPQA correlation drops slightly from the weak to the strong hint. Within the Qwen3-30B and Qwen3-80B families, hints reverse the ranking between thinking and instruct configurations: Qwen3-80B thinking trails its instruct counterpart without hints ($0.028$ versus $0.080$) but leads under the strong hint ($0.512$ versus $0.066$). Qwen3-30B shows the same reversal, and DeepSeek-V3.2 and GLM-4.7 likewise gain more in their reasoning-oriented modes.

Taken together, temporal-law hints reveal reasoning advantages hidden under the default instruction, which helps explain why strong reasoners can fall behind a smaller non-thinking model on TALD. This contrast suggests that the default preference for newer law can prevent models from making effective use of their existing reasoning capabilities. We next examine how the reasoning process changes when temporal-law hints make these capabilities more effective.

\subsection{Results on Hint-Elicited Exploration}
\label{sec:rq3}

% Queue the trend figure before the experimental design and findings.

The gains in RQ2 motivate RQ3: \textit{how do domain-specific hints elicit LLMs' latent TALD capabilities?} One possibility is that expert guidance makes models more confident in a preferred legal analysis. The newer-law bias suggests another: models may commit too readily to a familiar statutory version, and hints may help by reopening consideration of alternatives. Under this account, more accurate version selection should be accompanied by greater uncertainty during version-related reasoning.

\myparagraph{Experiment 6: Probing statutory-version exploration.}
% \label{ssec:rq3_hypothesis}
Prior work links token-level entropy to exploratory reasoning behavior~\citep{cheng2025reasoning, zhang-etal-2025-entropy}, so we use version-related policy entropy as a behavioral proxy of uncertainty during statutory-version reasoning.

\label{ssec:rq3_setup}

We identify \emph{version-related} positions in the model's CoT by matching law names and temporal-applicability expressions, such as retroactivity and effective periods. For each response $y_i$, let $S_i$ denote the selected token positions. At each position $t$, we compute:
\begin{equation*}
H_{i,t} = - \sum_{v \in \mathcal{V}}
p(v \mid y_{i,<t}, x_i)\log p(v \mid y_{i,<t}, x_i),
\end{equation*}
where $x_i$ is the case input under the given hint condition and $\mathcal{V}$ is the full vocabulary. The version-related entropy is:
\begin{equation*}
H_i^{ver} = \frac{1}{|S_i|}\sum_{t \in S_i} H_{i,t}.
\end{equation*}
Higher values indicate less concentrated next-token distributions at the selected positions, providing a local measure of uncertainty during version-related reasoning. We average $H_i^{ver}$ over old-law cases and compare hint conditions within each model (estimation details in Appendix~\ref{app:entropy}).

\label{sssec:rq3_probe_setup}
We retain experiment 5's task, prompts, and generation settings, locally delopyed the three Qwen thinking models and LegalOne-8B 
to obtain full-vocabulary probabilities for entropy estimation (Appendix~\ref{app:runs}). We report mean accuracy and entropy with standard errors over three sampling seeds.

\myparagraph{Finding 6: Higher TALD accuracy comes with greater version-related uncertainty (Figure~\ref{fig:rq3_main} and Table~\ref{tab:app_rq3_main}).}
\label{ssec:rq3_results}
For all four models, both hint conditions yield higher mean TALD accuracy and higher mean version-related policy entropy than the no-hint condition. For example, under the strong hint, Qwen3-30B improves from $0.020$ to $0.450$ in accuracy while its entropy rises from $0.315$ to $0.462$. Thus, the more accurate version selection comes with less concentrated predictions at version-related positions. However, accuracy and entropy do not always rise together as hints become stronger. For Qwen3-235B, entropy increases from $0.253$ under the weak hint to $0.266$ under the strong hint, while accuracy stays nearly unchanged ($0.524$ versus $0.523$). Higher entropy thus accompanies improvement over the default policy but does not guarantee further gains from stronger guidance.

This pattern is consistent with hints prompting models to reconsider version choices that their default policy treats as settled too readily. Although entropy alone cannot establish whether models explicitly consider more statutory alternatives, the observed changes provide behavioral evidence consistent with this interpretation. Together with RQ2, it points to a domain-specific gap between reasoning capability and the default TALD policy, echoing the capability--policy distinction studied in reasoning post-training~\citep{yue2025rlvr,wen2026rlvr}. These findings motivate reasoning post-training that encourages models to check temporal applicability and weigh competing statutory versions, so that stronger general reasoning translates into more reliable legal decisions.

\section{Related Work}

\noindent\textbf{Legal AI tasks.}
Legal AI research widely studies predicting legal authorities and outcomes from case facts, including judgment, charge, law-article, and citation prediction~\citep{luo2017learning, xiao2018cail, fei2023lawbench, han2026auslaw}. Legal fact prediction~\citep{liu2025legal} makes explicit one step these tasks take as given, deriving legal facts from litigants' evidence. TALD makes explicit another: these tasks label authorities statically, e.g., by article number, leaving implicit which version of the provision governs the case. TALD requires models to determine laws of versions with different effective periods, targeting a prerequisite of these tasks rather than a substitute for them.

\noindent\textbf{Temporal issues in legal AI.}
Prior work treats time as distribution shift in legal text \citep{santosh2024chronoslex, santosh2025lextempus}, as event ordering within legal narratives \citep{barale2025lextime}, or as statutory revisions to which judgment prediction must adapt \citep{han2026lawshift}, but does not ask which version should govern a given case. Closer to our work, LegalSearch-R1 \citep{fan2026timetravel} retrieves the provision valid at a reference time via temporally filtered search. TALD instead targets how that reference time is determined: in real judgment-derived disputes, models must identify the legally decisive events and apply temporal-effect rules, including exceptions favoring newer law, which date-filtered retrieval alone cannot resolve. We further diagnose failures by separating version bias, legal knowledge, and task-specific rule application.

% \myparagraph{LLM Reasoning}
% Reasoning-oriented post-training has improved LLM performance on tasks such as mathematics, logic, and programming~\citep{guo2025deepseek, xu2025toward}. Whether these capabilities transfer to tasks requiring domain-specific knowledge and judgment remains an important question~\citep{ma2026general, cheng2026revisiting}. TALD examines this issue through statutory version selection, comparing reasoning configurations and the effects of guidance about temporal applicability.

% Prior work connects predictive entropy with uncertainty and exploratory behavior during reasoning~\citep{cheng2025reasoning, zhang-etal-2025-entropy} and studies entropy dynamics during RL post-training~\citep{cui2025entropy}; LegalSearch-R1 also uses entropy-based advantage shaping during RL training \citep{fan2026timetravel}. We instead examine predictive entropy at version-related positions during inference, using it to characterize changes that accompany more accurate temporal-law judgments under guidance.

\section{Conclusion}

We proposed the Temporal Applicable-Law Determination task, and introduced TALD-Bench, a benchmark of Chinese civil judgments that evaluates whether LLMs determine the correct statutory version governing a case. Across nine models, strong accuracy on new-law cases masks severe failures on old-law cases, with errors systematically favoring newer law. Our diagnosis shows that this failure stems not from missing capabilities: models often recognize statutory versions and temporal-effect rules, stronger general reasoning does not reliably improve TALD, and even rule-free reminders yield substantial gains. These findings support a domain-specific reasoning-policy bottleneck in how existing capabilities are used. Rather than legal knowledge or general reasoning, future work should prioritize policy adaptation that elicits statutory alternatives and checks their temporal applicability. More broadly, separating limitations in capabilities from limitations in their use can turn benchmark failures into directions for specialized LLMs.

\section*{Limitations}
First, the scope of our dataset is limited to Chinese civil judgments, where our annotators can provide reliable legal expertise; although this setting contains rich statutory revisions and well represents the Civil Law Systems, our results may not directly generalize to other legal systems. 
Second, we study only the temporal axis of law version. Other applicability axes, such as jurisdictional version, likewise require selecting one provision among related candidates, and whether reasoning models exhibit analogous default-policy biases along them could be a meaningful direction for future research.
Third, our entropy analysis covers four open-weight models and characterizes changes under inference-time guidance. Establishing whether entropy mediates the accuracy gains, or whether RL training contributes to the default version bias, requires direct interventions beyond the hint comparisons studied here. Token entropy at version-related positions also does not directly measure the diversity of legal alternatives considered by a model.

\section*{Ethics Statement}

This work evaluates LLM behavior in legal reasoning tasks and should not be used as legal advice. Incorrect law-version determination can have serious consequences in real legal settings. We therefore emphasize that LLM outputs must be verified by qualified legal professionals. We collected judgments publicly released through China Judgments Online and used official statutory texts. These official judicial and legislative documents are excluded from copyright protection under Article 5 of the Copyright Law of the People's Republic of China. We preserve the anonymization present in the published judgments and rely on the courts' pre-publication processing. We do not conduct additional independent content screening. Under the applicable institutional rules, ethics review was not required for this study.

% References for the main text and supplementary methods.
\bibliography{custom}

\clearpage
\appendix
%\input{Sections/TALD_appendix_prefilled}
% Inline supplementary material, shared by main.tex and the appendix preview.
% All six tables and four figures are defined in this file.
% Include this file after the bibliography and the main file's appendix command.
% Select only one appendix version at a time: the versions share labels.
% Required packages: amsmath, amssymb, array, booktabs, multirow, hyperref,
% and natbib for the manuscript's citations; retain the main bibliography.
% BEGIN/END INLINE comments record the former source paths for navigation.
% Edit the blocks here: changes to the former source files do not update them.
% Original Chinese prompt strings: others/TALD_appendix_prompts_zh.json.
\providecommand{\TALDMainRef}[1]{\ref{#1}}

\makeatletter
\setlength{\@fptop}{0pt}
\setlength{\@fpsep}{12pt}
\setlength{\@dblfptop}{0pt}
\setlength{\@dblfpsep}{12pt}
\makeatother

\section{TALD-Bench Dataset}
\label{app:data}

\subsection{Case Selection and Quality Control}
The TALD-Bench dataset is drawn from civil judgments
issued between 2022 and 2024 and collected from China Judgments
Online in 2025. Using the standardized structure of Chinese civil
judgments, we apply regular-expression matching to extract case facts
and plaintiff claims as model inputs, and civil-law citations as
reference labels. The old-law and new-law cases are defined
by the statutory versions applied in the judgments.
The input excludes the court's reasoning, so models must determine
the applicable versions from the facts and claims. During collection, we controlled the ratio of old-law to new-law judgments to construct two equally sized splits.

Two legally trained coauthors checked a random sample of approximately 1\% of processed judgments for extraction accuracy. They divided the verification work, independently assessed their assigned materials, and then conducted a joint review. The Civil Code and applicable judicial interpretations served as legal references.

\subsection{Statutory Correspondences and Version Labels}
\label{app:version_labels}
Accurate interpretation of a model's statutory-version choices
requires relating the provisions it cites to their counterparts
under earlier or newer law. We establish article-level
correspondences using \emph{Civil Code of the People's Republic of
China: Comparisons with Prior and Related Provisions}, published by
China Legal Publishing House~\citep{civilcode2020comparison}.
The resulting dictionaries cover all 1,260 Code articles and link
them to the earlier provisions identified in the volume, where
available. The complete correspondences are provided in the
supplementary materials.

The version labels link earlier statutes to corresponding areas of
the Civil Code. For example, applying the Code's contract provisions
when the earlier Contract Law should govern is an old-to-new version
error. Table~\ref{tab:app_statute_families} provides the statute-level
mapping. Statute names and article numbers are normalized, and
repeated citations are removed when constructing the labels.

% BEGIN INLINE: tables/app_statute_families.tex
\begin{table*}[!tp]
\centering
\small
\setlength{\tabcolsep}{4pt}
\renewcommand{\arraystretch}{1.15}
\begin{tabular*}{\linewidth}{@{\extracolsep{\fill}}>{\raggedright\arraybackslash}p{0.21\linewidth}>{\raggedright\arraybackslash}p{0.39\linewidth}>{\raggedright\arraybackslash}p{\dimexpr0.40\linewidth-4\tabcolsep\relax}@{}}
\toprule
\textbf{Area of civil law} & \textbf{Predecessor statute(s)} & \textbf{Civil Code counterpart} \\
\midrule
General provisions & General Principles of the Civil Law; General Provisions of the Civil Law & Book I: General Provisions \\
\addlinespace[2pt]
Property & Property Law & Book II: Real Rights \\
\addlinespace[2pt]
Contract & Contract Law & Book III: Contracts \\
\addlinespace[2pt]
Personality rights & --- & Book IV: Personality Rights \\
\addlinespace[2pt]
Marriage and family & Marriage Law & Book V: Marriage and Family \\
\addlinespace[2pt]
Inheritance & Succession Law & Book VI: Succession \\
\addlinespace[2pt]
Tort liability & Tort Liability Law & Book VII: Tort Liability \\
\addlinespace[2pt]
Adoption & Adoption Law & Adoption provisions in Book V \\
\addlinespace[2pt]
Security & Security Law & Selected provisions in Books II and III \\
\bottomrule
\end{tabular*}
\caption{Predecessor statutes and their Civil Code counterparts,
grouped by area of civil law. A dash indicates that no separate
predecessor statute is listed for that area.}
\label{tab:app_statute_families}
\end{table*}
% END INLINE: tables/app_statute_families.tex

\section{Construction of the Knowledge Probes}
\label{app:knowledge}
The two probes in Section~\TALDMainRef{ssec:rq2_knowledge} assess complementary
components of legal knowledge: what each statutory version says,
and when each version applies.

\subsection{Statutory-Version Questions}
\label{app:memorization}

\myparagraph{Selection rationale.}
The Civil Code preserves much of the content of earlier statutes,
making substantive revisions particularly informative for testing
whether models distinguish their versions. Using the correspondences
in Appendix~\ref{app:version_labels}, the probe contrasts provisions
addressing the same legal issue but differing in substantive content.

\myparagraph{Expert selection of major revisions.}
Our legally trained annotators selected 100 provision pairs representing important
substantive changes, drawing on legal expertise and published
analyses of the reforms, including \citet{liu2020civilcode}.
Selection focuses on changes to legal conditions, rights and
obligations, or legal consequences. The resulting contrasts are
both identifiable from statutory content and relevant to legal
application. For example, Civil Code Article~1128 extends inheritance
by representation to children of a predeceased sibling, while
Article~11 of the earlier Succession Law covers descendants of a
predeceased child. The selected pairs span all seven books of the
Code.

\myparagraph{Question construction and evaluation.}
The fixed evaluation set contains 50 questions, sampled with seed~1
from the selected pairs. Each question uses one original statutory
text and asks which of two candidate statutes it comes from.
The set contains 25 earlier-statute provisions and 25 Civil Code
provisions, with one question per pair. When a revision adds
distinctive content to otherwise shared text, sampling favors the
extended Code provision while preserving this balance.

The alternative version's text is not shown. Option order is
shuffled once and held fixed across models, as are the selected
texts. Questions are answered independently, without retrieval or
preceding TALD context. We score the final option, counting
unanswered or unparseable responses as incorrect, and report
overall accuracy; random choice yields 0.50.
Prompt~\ref{prompt:app_version_probe} in Appendix~\ref{app:prompt_templates}
gives the prompt and an old-law question from the inheritance example above. The
supplementary materials provide the correspondence dictionaries,
the 100 pairs with selection rationales, and the complete Chinese
question set.

\subsection{Temporal-Effect Rule Questions}
\label{app:mcq}
Version knowledge alone does not determine which law governs a
case. This second probe tests knowledge of the temporal-effect
rules established by the Supreme People's Court's judicial
interpretation on applying the Civil Code~\citep{spc2020temporal}.
From an expert-designed pool of 32 questions, we use 16 covering
two aspects of that interpretation: eight on the general rules
(Articles~1--5) and eight on transitional application
(Articles~20--26). Both legal reviewers checked the selected
questions and reference answers.

Each question specifies a temporal condition and asks whether
the Civil Code or the relevant earlier law applies.
Prompt~\ref{prompt:app_probe_prompts} in Appendix~\ref{app:prompt_templates}
gives the common prompt and one example from each category. The general-rule example illustrates
non-retroactivity; the transitional example distinguishes the time
of contract formation from the time of performance. 
%The supplementary materials provide all 16 questions and reference answers in Chinese, together with English translations.

As in the statutory-version probe, questions are answered
independently without retrieval or prior case context. Accuracy is
the proportion of correct final options, with unparseable answers
counted as incorrect.

\section{Models and Evaluation Protocol}
\label{app:evals}
This section identifies the primary models and specifies generation,
sampling, error aggregation, and entropy estimation. Prompt conditions
are detailed in Appendix~\ref{app:prompts}; the full result tables are
organized by research question in Appendix~\ref{app:additional}.

\subsection{Model Names and Configurations}
\label{app:models}
Table~\ref{tab:app_model_ids} identifies the nine primary models used
in the RQ1--RQ3 analyses. The table maps these names to the API model IDs supplied
during evaluation or the Hugging Face repositories of the open-weight
models.
For Qwen, \textit{instruct} and \textit{thinking} select separate
checkpoints. For GLM, the same labels indicate whether thinking is
disabled or enabled. For GPT, Gemini, and Claude, the listed levels
are provider-specific inference-time reasoning settings.
The TALD accuracy tables specify the evaluated mode. Qwen and LegalOne
are run locally for RQ3 to obtain full-vocabulary token probabilities.

In our DeepSeek-V3.2 experiments, \texttt{deepseek-chat} and
\texttt{deepseek-reasoner} selected its non-thinking and thinking
modes, respectively. These are the API identifiers used at
evaluation time.\footnote{DeepSeek's official change log records
reassignment of these identifiers to V4-Flash on April~24, 2026,
and their scheduled discontinuation on July~24, 2026:
\url{https://api-docs.deepseek.com/updates/}.}

% BEGIN INLINE: tables/app_model_ids.tex
\begin{table*}[!t]
\centering
\footnotesize
\setlength{\tabcolsep}{4pt}
\renewcommand{\arraystretch}{1.15}
\begin{tabular*}{\linewidth}{@{\extracolsep{\fill}}>{\raggedright\arraybackslash}p{0.22\linewidth}>{\raggedright\arraybackslash}p{0.25\linewidth}>{\raggedright\arraybackslash}p{\dimexpr0.53\linewidth-4\tabcolsep\relax}@{}}
\toprule
\multicolumn{3}{l}{\textbf{(a) API model IDs used in evaluation}} \\
\midrule
\textbf{Name in paper} & \textbf{Mode(s) in results} & \textbf{API model ID} \\
\midrule
DeepSeek-V3.2 & chat & \texttt{deepseek-chat} \\
 & reasoner & \texttt{deepseek-reasoner} \\
\addlinespace[3pt]
GLM-4.7 & instruct; thinking & \texttt{z-ai/glm-4.7} \\
\addlinespace[3pt]
GPT-5.4 & none; low; medium; high & \texttt{gpt-5.4} \\
\addlinespace[3pt]
Gemini-3.1-Pro & low; medium; high & \texttt{gemini-3.1-pro-preview} \\
\addlinespace[3pt]
Claude Opus 4.6 & low; medium; high; max & \texttt{claude-opus-4-6} \\
\bottomrule
\end{tabular*}\par\vspace{8pt}
\begin{tabular*}{\linewidth}{@{\extracolsep{\fill}}>{\raggedright\arraybackslash}p{0.22\linewidth}>{\raggedright\arraybackslash}p{0.25\linewidth}>{\raggedright\arraybackslash}p{\dimexpr0.53\linewidth-4\tabcolsep\relax}@{}}
\toprule
\multicolumn{3}{l}{\textbf{(b) Open-weight model repositories}} \\
\midrule
\textbf{Name in paper} & \textbf{Mode in results} & \textbf{Hugging Face repository} \\
\midrule
Qwen3-235B & instruct & \href{https://huggingface.co/Qwen/Qwen3-235B-A22B-Instruct-2507}{\texttt{Qwen/Qwen3-235B-A22B-Instruct-2507}} \\
 & thinking & \href{https://huggingface.co/Qwen/Qwen3-235B-A22B-Thinking-2507}{\texttt{Qwen/Qwen3-235B-A22B-Thinking-2507}} \\
\addlinespace[3pt]
Qwen3-80B & instruct & \href{https://huggingface.co/Qwen/Qwen3-Next-80B-A3B-Instruct}{\texttt{Qwen/Qwen3-Next-80B-A3B-Instruct}} \\
 & thinking & \href{https://huggingface.co/Qwen/Qwen3-Next-80B-A3B-Thinking}{\texttt{Qwen/Qwen3-Next-80B-A3B-Thinking}} \\
\addlinespace[3pt]
Qwen3-30B & instruct & \href{https://huggingface.co/Qwen/Qwen3-30B-A3B-Instruct-2507}{\texttt{Qwen/Qwen3-30B-A3B-Instruct-2507}} \\
 & thinking & \href{https://huggingface.co/Qwen/Qwen3-30B-A3B-Thinking-2507}{\texttt{Qwen/Qwen3-30B-A3B-Thinking-2507}} \\
\addlinespace[3pt]
LegalOne-8B & reasoning & \href{https://huggingface.co/CSHaitao/LegalOne-8B}{\texttt{CSHaitao/LegalOne-8B}} \\
\bottomrule
\end{tabular*}
\caption{Identifiers and evaluated configurations of the nine primary models. Panel (a) reports API request identifiers used
during evaluation; panel (b) links to the corresponding Hugging Face
model repositories.}
\label{tab:app_model_ids}
\label{tab:app_qwen_ids}
\end{table*}
% END INLINE: tables/app_model_ids.tex

\subsection{Generation, Sampling, and Aggregation}
\label{app:runs}
We set temperature and top-$p$ to 1.0. Output limits are set
separately for each model to allow complete responses. The
no-hint, weak-hint, and strong-hint conditions use the same case
inputs and output format; Appendix~\ref{app:prompts} describes
the guidance supplied in each condition.

Each two-split TALD run samples 100 cases from each split. The
hint comparisons and entropy analysis use 100 old-law cases
per run. Within a run, models and prompt conditions share the
sampled cases. Sampling seeds vary between runs, which can contain
overlapping cases. API-based TALD accuracy in RQ1 and RQ2 is
averaged over four sampling seeds. RQ3 uses three seeds for local
inference and entropy estimation.

TALD accuracy is averaged first over cases and then over runs.
Standard errors are computed across runs: four for the primary API
evaluations and three for RQ3.
The knowledge probes use the fixed 50- and 16-question sets
described in Appendix~\ref{app:knowledge}, shared across models. 

For Qwen3, RQ1 and RQ2 use API-served inference, while RQ3 uses local deployment to access full-vocabulary probabilities. The task, prompts, and generation hyperparameters remain unchanged, but the obtained performance can differ between API and local runs. Figure~\ref{fig:rq3_main} and Table~~\ref{tab:app_rq3_mainv} report accuracy and entropy from the local deployments for RQ3.

The local hardware configuration comprised eight NVIDIA H200 GPUs. Local generation uses vLLM. TALD scoring is implemented using custom Python scripts, while teacher-forced token distributions are obtained using Hugging Face Transformers with PyTorch. Entropy is computed in float32 over the full vocabulary. Software package versions were not consistently pinned across runs as the available computational resources changed during the study.

\subsection{Error-Type Analysis}
\label{app:fault_types}
A case may require several laws. We examine each required version
separately, recording an error if it is not correctly identified.
A missed earlier-law version is \textsc{Old-to-New} when the
response cites any Civil Code target version; a missed Civil Code
version is \textsc{New-to-Old} when an earlier-law target version
is cited. \textsc{Irrelevant-Law} label the type of error when any version of the required law could not be predicted by the LLM. Citing both versions of the same statutory family is treated as an incorrect prediction. Such a conflict is classified as OLD-TO-NEW when the reference requires the earlier version, and NEW-TO-OLD when it requires the Civil Code version.

For a case requiring $K$ laws, each missed version contributes an
error weight of $1/K$. Thus, two \textsc{Old-to-New} errors in a
case requiring three laws contribute $2/3$ to that category.
For each model, we pool these weights across both splits and
divide each category's weight by the total error weight.
Figure~\TALDMainRef{fig:fault_types} reports these weighted error
proportions, which sum to one.

\subsection{Version-Related Entropy Estimation}
\label{app:entropy}
RQ3 focuses on uncertainty at reasoning positions that concern
legal versions. We identify these positions by matching law names
and temporal-applicability expressions in the CoT, including the
Civil Code, Contract Law, legal effect, and entry into force.
The matched positions form the set $S_i$ used in
Section~\TALDMainRef{ssec:rq3_setup}.

For each generated CoT, we teacher-force its tokens through the
model to obtain the conditional next-token distributions.
Entropy is computed over the full vocabulary with natural
logarithms, averaged over the matched positions within each
case, and then averaged over cases. The same procedure is used
for all hint conditions within a model. The statistic summarizes local
next-token uncertainty at version-related positions.

The complete local-inference results are reported in
Appendix~\ref{app:rq3_full_results}.

\section{Temporal-Law Guidance}
\label{app:prompts}
The hint intervention in Section~\TALDMainRef{sssec:rq2_hint} varies
the system instruction while keeping the case input and requested output
format fixed. Appendix~\ref{app:prompt_templates} gives the baseline
system instruction and shared user prompt in
Prompt~\ref{prompt:app_default_prompt}, and the two alternative system
instructions in Prompts~\ref{prompt:app_weak_hint}
and~\ref{prompt:app_strong_hint}. No external retrieval
is used.

\myparagraph{No hint.}
The baseline uses a generic assistant instruction and asks which legal
provisions apply, without directing attention to temporal applicability.

\myparagraph{Weak hint.}
The system instruction adds a reminder to consider event timing,
temporal effect, and retroactivity. It does not state a legal rule for
choosing between old and new law or provide a case-specific answer.

\myparagraph{Strong hint.}
The strong hint retains this reminder and adds a brief doctrinal
summary. Our legal experts grounded it in Articles~1--4 of the Supreme
People's Court's temporal-effect interpretation~\citep{spc2020temporal}
and its official explanation~\citep{yang2020temporal}: non-retroactivity
(Article~1), beneficial retroactivity (Article~2), and gap-filling
application of new provisions (Article~3). Article~4 distinguishes
using the Code's more specific provisions to inform judicial reasoning
from replacing the earlier law that governs the dispute.

\myparagraph{Relation to the knowledge probe.}
These principles are also examined by the temporal-effect rule probe
in Appendix~\ref{app:mcq}; its first seven questions address
Articles~1--4. The hints bring this knowledge into case analysis,
varying how explicitly the relevant rules are stated. The supplementary
materials link each principle to its probe questions.

\section{Additional Evaluation Results}
\label{app:additional}
We organize the detailed results by research question: versional
performance differences in RQ1, alignment with external reasoning
benchmarks in RQ2, and changes in version-related uncertainty in RQ3.

\subsection{RQ1: Performance Across Statutory Versions}
\label{app:additional_r1}
% BEGIN INLINE: tables/app_rq1_full.tex
\begin{table*}[!t]
\centering
\begingroup
\small
\setlength{\tabcolsep}{4pt}
\renewcommand{\arraystretch}{1.15}
\begin{tabular*}{\textwidth}{@{\extracolsep{\fill}}llcc@{}}
\toprule
\multirow{2}{*}{\textbf{Model}} & \multirow{2}{*}{\textbf{Mode}} & \textbf{Old-law} & \textbf{New-law} \\
\cmidrule(lr){3-3} \cmidrule(lr){4-4}
 & & ACC $\pm$ SE & ACC $\pm$ SE \\
\midrule
DeepSeek-V3.2 & reasoner & $0.092\,\pm\,0.0167$ & $0.746\,\pm\,0.0155$ \\
GLM-4.7 & thinking & $0.080\,\pm\,0.0240$ & $0.787\,\pm\,0.0149$ \\
GPT-5.4 & high & $0.237\,\pm\,0.0048$ & $0.701\,\pm\,0.0070$ \\
Gemini-3.1-Pro & high & $0.135\,\pm\,0.0130$ & $0.843\,\pm\,0.0198$ \\
Claude Opus 4.6 & max & $0.014\,\pm\,0.0040$ & $0.820\,\pm\,0.0198$ \\
Qwen3-235B & thinking & $0.148\,\pm\,0.0170$ & $0.730\,\pm\,0.0288$ \\
Qwen3-80B & thinking & $0.028\,\pm\,0.0050$ & $0.798\,\pm\,0.0131$ \\
Qwen3-30B & thinking & $0.020\,\pm\,0.0075$ & $0.765\,\pm\,0.0145$ \\
LegalOne-8B & reasoning & $0.093\,\pm\,0.0078$ & $0.800\,\pm\,0.0138$ \\
\bottomrule
\end{tabular*}
\endgroup
\caption{Default-instruction TALD accuracy for the nine primary models, corresponding to Figure~\TALDMainRef{fig:rq1_initial}. Each entry reports the mean $\pm$ its standard error across four sampling seeds; each run samples 100 cases from each split.}
\label{tab:tald_old_new_se}
\end{table*}
% END INLINE: tables/app_rq1_full.tex

Table~\ref{tab:tald_old_new_se} gives the complete values underlying
the primary RQ1 comparison in Figure~\TALDMainRef{fig:rq1_initial}.
It reports the same nine model configurations under the default
instruction, with mean accuracy and standard errors for both splits.
All nine achieve above $0.70$ on new-law cases and below $0.25$ on
old-law cases. The paired columns make the magnitude of the versional
gap explicit for each model.

\myparagraph{Supplementary evaluation of newer frontier models.}
\label{app:additional_api}
% BEGIN INLINE: tables/app_additional_api.tex
% Original means and standard errors are preserved.
\begin{table*}[!t]
\centering
\begingroup
\small
\setlength{\tabcolsep}{3pt}
\renewcommand{\arraystretch}{1.15}
\newcommand{\taldapivalue}[2]{\ensuremath{#1}\,{\scriptsize\ensuremath{\pm\,#2}}}
\begin{minipage}[t]{0.485\textwidth}
\vspace{0pt}
\begin{tabular*}{\linewidth}{@{\extracolsep{\fill}}lcc@{}}
\toprule
\textbf{Mode} & \textbf{Old-law} & \textbf{New-law} \\
\midrule
\multicolumn{3}{@{}l}{\textbf{GPT-5.6 Sol}} \\
\addlinespace[1pt]
none & \taldapivalue{0.124}{0.0153} & \taldapivalue{0.788}{0.0147} \\
low & \taldapivalue{0.406}{0.0164} & \taldapivalue{0.688}{0.0310} \\
medium & \taldapivalue{0.410}{0.0102} & \taldapivalue{0.682}{0.0246} \\
high & \taldapivalue{0.366}{0.0130} & \taldapivalue{0.710}{0.0118} \\
xhigh & \taldapivalue{0.355}{0.0098} & \taldapivalue{0.710}{0.0292} \\
max & \taldapivalue{0.329}{0.0290} & \taldapivalue{0.743}{0.0192} \\
\specialrule{0.25pt}{3pt}{3pt}
\multicolumn{3}{@{}l}{\textbf{GPT-5.6 Terra}} \\
\addlinespace[1pt]
none & \taldapivalue{0.014}{0.0073} & \taldapivalue{0.765}{0.0167} \\
low & \taldapivalue{0.211}{0.0114} & \taldapivalue{0.759}{0.0146} \\
medium & \taldapivalue{0.207}{0.0153} & \taldapivalue{0.744}{0.0338} \\
high & \taldapivalue{0.250}{0.0181} & \taldapivalue{0.739}{0.0233} \\
xhigh & \taldapivalue{0.243}{0.0069} & \taldapivalue{0.738}{0.0094} \\
max & \taldapivalue{0.202}{0.0122} & \taldapivalue{0.787}{0.0211} \\
\specialrule{0.25pt}{3pt}{3pt}
\multicolumn{3}{@{}l}{\textbf{GPT-5.6 Luna}} \\
\addlinespace[1pt]
none & \taldapivalue{0.029}{0.0064} & \taldapivalue{0.803}{0.0282} \\
low & \taldapivalue{0.199}{0.0169} & \taldapivalue{0.666}{0.0102} \\
medium & \taldapivalue{0.272}{0.0168} & \taldapivalue{0.667}{0.0297} \\
high & \taldapivalue{0.333}{0.0129} & \taldapivalue{0.596}{0.0056} \\
xhigh & \taldapivalue{0.360}{0.0147} & \taldapivalue{0.614}{0.0075} \\
max & \taldapivalue{0.337}{0.0171} & \taldapivalue{0.651}{0.0173} \\
\bottomrule
\end{tabular*}
\end{minipage}
\hfill
\begin{minipage}[t]{0.485\textwidth}
\vspace{0pt}
\begin{tabular*}{\linewidth}{@{\extracolsep{\fill}}lcc@{}}
\toprule
\textbf{Mode} & \textbf{Old-law} & \textbf{New-law} \\
\midrule
\multicolumn{3}{@{}l}{\textbf{Claude Opus 5.5}} \\
\addlinespace[1pt]
low & \taldapivalue{0.314}{0.0325} & \taldapivalue{0.703}{0.0159} \\
medium & \taldapivalue{0.314}{0.0123} & \taldapivalue{0.689}{0.0182} \\
high & \taldapivalue{0.279}{0.0265} & \taldapivalue{0.678}{0.0326} \\
xhigh & \taldapivalue{0.310}{0.0128} & \taldapivalue{0.673}{0.0223} \\
max & \taldapivalue{0.323}{0.0365} & \taldapivalue{0.717}{0.0064} \\
\specialrule{0.25pt}{3pt}{3pt}
\multicolumn{3}{@{}l}{\textbf{Claude Sonnet 5.5}} \\
\addlinespace[1pt]
low & \taldapivalue{0.043}{0.0144} & \taldapivalue{0.768}{0.0257} \\
medium & \taldapivalue{0.042}{0.0112} & \taldapivalue{0.766}{0.0144} \\
high & \taldapivalue{0.248}{0.0168} & \taldapivalue{0.705}{0.0320} \\
xhigh & \taldapivalue{0.269}{0.0251} & \taldapivalue{0.699}{0.0185} \\
max & \taldapivalue{0.125}{0.0194} & \taldapivalue{0.727}{0.0208} \\
\specialrule{0.25pt}{3pt}{3pt}
\multicolumn{3}{@{}l}{\textbf{Claude Haiku 5.5}} \\
\addlinespace[1pt]
low & \taldapivalue{0.034}{0.0151} & \taldapivalue{0.754}{0.0158} \\
medium & \taldapivalue{0.053}{0.0100} & \taldapivalue{0.734}{0.0160} \\
high & \taldapivalue{0.068}{0.0136} & \taldapivalue{0.722}{0.0220} \\
xhigh & \taldapivalue{0.094}{0.0029} & \taldapivalue{0.719}{0.0023} \\
max & \taldapivalue{0.005}{0.0028} & \taldapivalue{0.314}{0.0266} \\
\bottomrule
\end{tabular*}
\end{minipage}
\endgroup
\caption{Supplementary evaluation of newer frontier models under the default instruction. These configurations assess whether the old-law performance deficit persists beyond the primary model panel and are not included in the knowledge, hint, or entropy analyses. Entries report mean TALD accuracy and its across-seed standard error.}
\label{tab:tald_api_reasoning_mean_se}
\end{table*}
% END INLINE: tables/app_additional_api.tex

To assess whether TALD remains challenging for newer frontier models, we additionally evaluate the GPT-5.6 and Claude 5.5 families under the default instruction. These supplementary evaluations are separate from the primary model panel used for the diagnostic analyses. Every evaluated configuration retains an old-law performance deficit relative to new-law cases, indicating that TALD remains unresolved under this evaluation protocol. Performance is reported in Table~\ref{tab:tald_api_reasoning_mean_se}. Every listed configuration performs better on new-law
than on old-law cases. Higher effort does not uniformly improve
old-law accuracy within a family: for example, GPT-5.6 Sol reports
$0.406$ at low effort and $0.329$ at max effort. These supplementary
evaluations show that the versional gap persists beyond the primary
model panel; the knowledge and hint diagnoses in RQ2 and RQ3 focus on
the primary models described in Appendix~\ref{app:models}.

\subsection{RQ2: External General Reasoning Benchmarks and Hint Alignment}
\label{app:reasoning_benchmarks}
% BEGIN INLINE: tables/app_external_reasoning_v2.tex
% Inline snapshot of the external-score table; its separate generation script
% does not update this embedded copy.
\begin{table*}[!t]
\centering
\begingroup\footnotesize
\setlength{\tabcolsep}{3pt}
\renewcommand{\arraystretch}{1.15}
\begin{tabular*}{\textwidth}{@{\extracolsep{\fill}}llrrrrrr@{}}
\toprule
\multirow{2}{*}{\textbf{Model}} & \multirow{2}{*}{\textbf{Mode}} & \multicolumn{3}{c}{\textbf{External benchmark score}} & \multicolumn{3}{c}{\textbf{Old-law TALD accuracy}} \\
\cmidrule(lr){3-5}\cmidrule(lr){6-8}
 & & \textbf{GPQA} & \textbf{HLE} & \textbf{LCB} & \textbf{No hint} & \textbf{Weak hint} & \textbf{Strong hint} \\
\midrule
DeepSeek-V3.2 & chat & 75.1 & 11.2 & 59.3 & 0.013 & 0.028 & 0.080 \\
 & reasoner & 84.0 & 24.6 & 86.2 & 0.092 & 0.394 & 0.436 \\
\addlinespace[3pt]
Claude Opus 4.6 & max & 89.6 & 39.9 & -- & 0.014 & 0.113 & 0.156 \\
\addlinespace[3pt]
GPT-5.4 & none & 74.8 & 11.3 & -- & 0.021 & 0.087 & 0.167 \\
 & low & 87.1 & 30.8 & -- & 0.082 & 0.444 & 0.456 \\
\addlinespace[3pt]
GLM-4.7 & instruct & 66.4 & 6.4 & 56.2 & 0.013 & 0.032 & 0.053 \\
 & thinking & 85.9 & 27.4 & 89.4 & 0.080 & 0.337 & 0.435 \\
\addlinespace[3pt]
Qwen3-235B & instruct & 75.3 & 11.1 & 52.4 & 0.035 & 0.035 & 0.041 \\
 & thinking & 79.0 & 15.9 & 78.8 & 0.148 & 0.330 & 0.323 \\
\addlinespace[3pt]
Qwen3-80B & instruct & 73.8 & 7.6 & 68.4 & 0.080 & 0.078 & 0.066 \\
 & thinking & 75.9 & 12.6 & 78.4 & 0.028 & 0.386 & 0.512 \\
\addlinespace[3pt]
Qwen3-30B & instruct & 65.9 & 6.9 & 51.5 & 0.292 & 0.264 & 0.164 \\
 & thinking & 70.7 & 10.3 & 70.7 & 0.020 & 0.350 & 0.395 \\
\midrule
\multicolumn{2}{l}{Matched $N$} & 13 & 13 & 10 &  &  &  \\
\multicolumn{2}{l}{Spearman $\rho$: No hint} & 0.146 & 0.163 & 0.189 &  &  &  \\
\multicolumn{2}{l}{Spearman $\rho$: Weak hint} & 0.467 & 0.549 & 0.697 &  &  &  \\
\multicolumn{2}{l}{Spearman $\rho$: Strong hint} & 0.456 & 0.577 & 0.758 &  &  &  \\
\bottomrule
\end{tabular*}
\endgroup
\caption{Artificial Analysis benchmark scores (0--100) and old-law TALD accuracy (0--1) under no, weak, and strong hints. TALD columns report the across-seed means from Table~\TALDMainRef{tab:default_reasoning}. The bottom rows give each benchmark's matched configuration count and Spearman correlation with TALD under each hint condition; matching is fixed across conditions. Dashes indicate unavailable scores.}
\label{tab:app_external_reasoning}
\end{table*}
% END INLINE: tables/app_external_reasoning_v2.tex

% Queue the final wide table early enough to share the last results page.
% BEGIN INLINE: tables/rq3_main.tex
\begin{table*}[!t]
\centering
\begingroup
% ACL defines small and footnotesize identically; set the larger size explicitly.
\fontsize{10}{11}\selectfont
\setlength{\tabcolsep}{4pt}
\renewcommand{\arraystretch}{1.55}
% Keep the Table 1 arrangement: a full-size mean above a smaller standard error.
\newcommand{\taldresult}[2]{%
  \begingroup
  \renewcommand{\arraystretch}{0.8}%
  \begin{tabular}[c]{@{}c@{}}%
    #1\\[-2pt]%
    {\scriptsize\textpm{}\,#2}%
  \end{tabular}%
  \endgroup
}
\begin{tabular*}{\linewidth}{@{\extracolsep{\fill}}llcccccc@{}}
\toprule
\multirow{2}{*}[-4pt]{\textbf{Model}} & \multirow{2}{*}[-4pt]{\textbf{Mode}}
& \multicolumn{2}{c}{\textbf{No hint}}
& \multicolumn{2}{c}{\textbf{Weak hint}}
& \multicolumn{2}{c}{\textbf{Strong hint}} \\
\cmidrule(lr){3-4}\cmidrule(lr){5-6}\cmidrule(lr){7-8}
& & ACC & $H_i^{ver}$ & ACC & $H_i^{ver}$ & ACC & $H_i^{ver}$ \\
\midrule
Qwen3-235B  & thinking  & \taldresult{0.155}{0.025} & \taldresult{0.238}{0.006} & \taldresult{0.524}{0.025} & \taldresult{0.253}{0.004} & \taldresult{0.523}{0.032} & \taldresult{0.266}{0.005} \\
\specialrule{0.25pt}{1.125pt}{1.125pt}
Qwen3-80B   & thinking  & \taldresult{0.050}{0.002} & \taldresult{1.148}{0.026} & \taldresult{0.459}{0.022} & \taldresult{1.185}{0.029} & \taldresult{0.578}{0.024} & \taldresult{1.211}{0.025} \\
\specialrule{0.25pt}{1.125pt}{1.125pt}
Qwen3-30B   & thinking  & \taldresult{0.020}{0.015} & \taldresult{0.315}{0.005} & \taldresult{0.371}{0.015} & \taldresult{0.425}{0.010} & \taldresult{0.450}{0.020} & \taldresult{0.462}{0.014} \\
\specialrule{0.25pt}{1.125pt}{1.125pt}
LegalOne-8B & reasoning & \taldresult{0.093}{0.008} & \taldresult{0.147}{0.006} & \taldresult{0.280}{0.032} & \taldresult{0.159}{0.002} & \taldresult{0.298}{0.011} & \taldresult{0.173}{0.003} \\
\bottomrule
\end{tabular*}
\endgroup
\caption{
TALD accuracy and version-related policy entropy $H_i^{ver}$ on old-law cases across hint strength. Each cell shows the mean above and $\pm$ its standard error below, computed over three random seeds.
}
\label{tab:app_rq3_main}
% Preserve compatibility with the label used in the current manuscript.
\label{tab:app_rq3_mainv}
\end{table*}
% END INLINE: tables/rq3_main.tex

Table~\ref{tab:app_external_reasoning} provides the matched scores
and correlations underlying the no-hint, weak-hint, and strong-hint
heatmaps in Figure~\TALDMainRef{fig:reasoning_correlations}a--c.
Its upper rows pair external benchmark scores with old-law TALD
accuracy; its lower rows report sample sizes and correlations with TALD.
Only TALD scores vary across panels; external benchmark correlations
remain fixed.

\myparagraph{Benchmark coverage.}
GPQA Diamond~\citep{rein2024gpqa} tests graduate-level scientific
knowledge and reasoning through multiple-choice questions in biology,
chemistry, and physics. Humanity's Last Exam
(HLE)~\citep{hle2026expertlevel} tests expert-level academic problem
solving across disciplines; we use Artificial Analysis's text-only
evaluation. LiveCodeBench (LCB)~\citep{jain2024livecodebench} tests
algorithmic programming through competitive programming problems,
evaluated by Artificial Analysis using Python and pass@1. These three
benchmarks provide complementary reasoning measures, with GPQA and
HLE also depending on subject knowledge.

\myparagraph{Scores and configuration matching.}
External scores come from Artificial
Analysis~\citep{artificialanalysis2026methodology}, retrieved on
October~2, 2026. Each observation is a matched model release or
checkpoint and reasoning mode or effort level. There are 13 matched
configurations across seven of the nine primary models for GPQA and
HLE, and ten for LCB. The Gemini record does not resolve the required
effort setting, and LegalOne-8B has no matched record. A dash denotes
an unavailable score, not zero. Artificial Analysis uses its own
inference protocol; matching the release and mode does not imply an
identical API snapshot or evaluation setup.

\myparagraph{Correlation analysis.}
For each benchmark, we compute Spearman's $\rho$ across configurations
with both scores available. Each benchmark uses the same matched
configurations across hint conditions; correlations between external
benchmarks use their shared configurations. The footer reports each
sample size and its three hint-conditioned correlations. Both hints
improve alignment with TALD over no hint for all three benchmarks.
GPQA decreases slightly from weak to strong, so alignment does not
increase monotonically with hint strength in every benchmark.

\subsection{RQ3: Accuracy and Version-Related Entropy}
\label{app:rq3_full_results}
Table~\ref{tab:app_rq3_main} reports the absolute values underlying
Figure~\TALDMainRef{fig:rq3_main}, pairing TALD accuracy with
version-related policy entropy. Means and standard errors use three
sampling seeds under the local protocol in Appendix~\ref{app:runs},
with entropy estimated as described in Appendix~\ref{app:entropy}.

Within every model, both hints raise mean accuracy and entropy over
no hint. Qwen3-235B's entropy also rises from weak to strong hints,
yet accuracy barely changes. Higher entropy need not mean higher
accuracy.

\FloatBarrier
\section{Prompt Templates}
\label{app:prompt_templates}
The following prompts give English translations of the Chinese prompt
templates used in the experiments. The original templates are included
in the supplementary materials. Prompts~\ref{prompt:app_version_probe}
and~\ref{prompt:app_probe_prompts} give the statutory-version and
temporal-effect rule probes, respectively, with example questions.
Prompt~\ref{prompt:app_default_prompt} gives the no-hint system instruction
and the user prompt shared across all TALD hint conditions;
Prompts~\ref{prompt:app_weak_hint} and~\ref{prompt:app_strong_hint}
give the weak- and strong-hint system instructions, respectively. Only the system instruction changes across hint conditions.

% BEGIN INLINE: figures/app_version_probe.tex
\begin{figure*}[!t]
\centering
\tcbset{promptbox/.append style={breakable=false}}
\begin{prompt}{Statutory-version knowledge probe}
\label{prompt:app_version_probe}
\small
\textit{System.} You are taking a multiple-choice test on knowledge
of old and new versions of Chinese substantive civil law. You will
see the original text of a legal provision and two possible statutory
versions. Identify which version the provision comes from. Your
final answer must contain only one available uppercase option letter,
such as A or B. Do not explain your reasoning or output any other text.
\medskip\par
\textit{User.} Below is the original text of a provision of Chinese
substantive civil law. Identify which statutory version it comes from.\par
Provision: \texttt{\{article\_text\}}\par
Options: A. \texttt{\{choices[A]\}}; B. \texttt{\{choices[B]\}}\par
Output only the uppercase letter of the correct option, such as A
or B. Do not explain.
\medskip\hrule\medskip
\textbf{Example from the fixed question set}\par
\textit{Provision.} Where a child of the deceased dies before the
deceased, the child's direct lineal descendants inherit by
representation. A successor by representation generally inherits
only the share to which their father or mother would have been entitled.\par
\textit{Options.} A. Civil Code of the People's Republic of China,
Book VI: Succession; B. Succession Law of the People's Republic of China.\par
\textit{Correct answer.} B.
\end{prompt}
{\small\noindent English translations of the statutory-version prompt and
one old-law question. The answer is shown for illustration and is
not included in the model input.}
\end{figure*}
% END INLINE: figures/app_version_probe.tex

% BEGIN INLINE: figures/app_probe_prompts.tex
\begin{figure*}[!tp]
\centering
\tcbset{promptbox/.append style={breakable=false}}
\begin{prompt}{Temporal-effect rule probe}
\label{prompt:app_probe_prompts}
\small
\textbf{System: Temporal-effect rule probe}\par
You are taking a single-choice test on the temporal-effect rules of
Chinese civil substantive law. Based on the question, recall the
general principles governing the temporal effect of civil law and
the Provisions of the Supreme People's Court on the Temporal Effect
of the Application of the Civil Code of the People's Republic of
China, and choose the most appropriate answer. Your final response
must contain only one available uppercase option letter, such as
A or B. Do not explain your reasoning or output any other text.
\medskip\par
\textbf{User: Temporal-effect rule probe}\par
Condition: \texttt{\{condition\}}\par
Question: \texttt{\{question\}}\par
Options:\par
A. \texttt{\{choices[A]\}}\par
B. \texttt{\{choices[B]\}}\par
Output only the letter of the correct option.

\medskip\hrule\medskip
\textbf{Examples from the fixed question set}\par
\textit{Question (both examples).} Which rules of substantive civil
law should apply?\par
\textit{Options (both examples).} A. The Civil Code or its relevant
provisions (new law); B. The laws and judicial interpretations in
force at the relevant earlier time (old law).
\smallskip\par
\textit{General rule (Article~1(2)).} A civil dispute arises from
legal facts occurring before the Civil Code takes effect, with no
exception specified by a law or judicial interpretation.\par
\textit{Correct answer.} B.
\smallskip\par
\textit{Transitional rule (Article~20).} A contract was formed before
the Civil Code took effect, and performance continues afterwards by
law or agreement. The dispute concerns performance occurring after
the Code took effect.\par
\textit{Correct answer.} A.
\end{prompt}
{\small\noindent English translations of the temporal-effect rule prompt
and two example questions. Source provisions refer to the judicial
interpretation. Answers and source-article references are shown for
illustration and are not included in the model input.}
\end{figure*}
% END INLINE: figures/app_probe_prompts.tex

\clearpage
% BEGIN INLINE: figures/app_default_prompt.tex
\begin{figure*}[!tp]
\centering
\tcbset{promptbox/.append style={breakable=false}}
\begin{prompt}{TALD: No Hint}
\label{prompt:app_default_prompt}
If you encounter the following legal case, which legal provisions
should apply? Please respond in the format ``The applicable provisions
are Articles X, X, X of [XXX]; Articles Y, Y, Y of [YYY]; and Articles
O, O, O of [OOO].'' Write the full names of the laws, such as the
[XXX] Law of the People's Republic of China. The case is as follows:
\texttt{\{case\_facts\}\{plaintiff\_claims\}}
\end{prompt}
{\small\noindent TALD prompt under the no-hint condition.}
\end{figure*}
% END INLINE: figures/app_default_prompt.tex

% BEGIN INLINE: figures/app_hint_prompts.tex
\begin{figure*}[!tp]
\centering
\tcbset{promptbox/.append style={breakable=false}}
\begin{prompt}{TALD: Weak Hint}
\label{prompt:app_weak_hint}
Users may ask you legal consultation
questions; please try to answer them factually. Pay attention to the
timing of events in the case and to the temporal effect of the law.
Pay attention to issues of retroactivity.
\end{prompt}
{\small\noindent English translation of the weak-hint system instruction.}
\end{figure*}

\begin{figure*}[!tp]
\centering
\tcbset{promptbox/.append style={breakable=false}}
\begin{prompt}{TALD: Strong Hint}
\label{prompt:app_strong_hint}
\textbf{System: Strong temporal-law hint}\par
You are a helpful assistant. Users may ask you legal consultation
questions; please try to answer them factually. Pay attention to the
timing of events in the case and to the temporal effect of the law.
Pay attention to issues of retroactivity. The basic principle governing
the temporal effect of civil law is non-retroactivity, so particular
attention should be paid to whether the timing of the case facts
requires applying the old law. There are, however, exceptions, such as
beneficial retroactivity (where this better protects the lawful rights
and interests of civil subjects), retroactivity of newly introduced
provisions (where the old law had no provision and the new law adds
one), and circumstances in which newly introduced specific provisions
may be used in judicial reasoning.
\end{prompt}
{\small\noindent English translation of the strong-hint system instruction.}
\end{figure*}
% END INLINE: figures/app_hint_prompts.tex

\end{document}